\PassOptionsToPackage{table}{xcolor}
\documentclass[sigconf,nonacm]{acmart}
\usepackage{graphicx}   % \resizebox
\usepackage{booktabs}   % \cmidrule, \cmidrulewidth
\usepackage{multirow}   % \multirow
\usepackage{xcolor} % \cellcolor, \textcolor（table option passed above）
\usepackage{algorithm}
\usepackage{algorithmic}
\AtBeginDocument{%
  }

\setcopyright{none}

\begin{document}

\newcommand{\gformat}[2]{$\mathop{{{\textbf{#1}}}}\limits_{\textcolor[rgb]{0,0.68,0.02}
{\textbf{#2}}}$}
\newcommand{\red}[1]{\textcolor{cyan}{#1}}
\hypersetup{
    colorlinks=true,
    linkcolor=red,
    citecolor=cyan,
    filecolor=magenta,      
    urlcolor=cyan,
    }

\title{MindAdapter: Few-Shot Parameter-Efficient Residual Calibration of Cross-Subject Brain-to-Visual Decoding Models}

% \affiliation{%

\author{Jiaxiang Liu}
\affiliation{%
  \institution{Guangdong Institute of Intelligence Science and Technology}
  \city{Hengqin}
  \country{China}}
\email{liujiaxiang@gdiist.cn}

\author{Jiawei Du}
\affiliation{%
  \institution{Agency for Science, Technology and Research}
  % \city{Singapore}
  \country{Singapore}}
\email{dujiawei@u.nus.edu}

\author{Xupeng Chen}
\affiliation{%
  \institution{New York University}
  \city{Brooklyn}
  \country{US}}
\email{xc1490@nyu.edu}

\author{Guoqi Li}
\affiliation{%
  \institution{Institute of Automation, Chinese Academy of Sciences}
  \city{Beijing}
  \country{China}}
\email{guoqi.li@ia.ac.cn}

\author{Jiang Cai}
\affiliation{%
  \institution{Guangdong Institute of Intelligence Science and Technology}
  \city{Hengqin}
  \country{China}}
\email{caijiang@gdiist.cn}

\author{Simon Fong}
\affiliation{%
  \institution{Department of
Computer and Information Science, University of Macau}
  \city{Macau}
  \country{China}}
\email{ccfong@um.edu.mo}

\author{Mingkun Xu}
\affiliation{%
  \institution{Guangdong Institute of Intelligence Science and Technology}
  \city{Hengqin}
  \country{China}}
\email{xumingkun@gdiist.cn}

\begin{abstract}
Cross-subject brain-to-visual decoding remains a core challenge in brain–computer interfaces due to severe inter-individual variability that induces systematic subject-specific functional misalignment. To address this issue, we propose \textit{MindAdapter}, a parameter-efficient few-shot calibration framework for pretrained brain-to-visual decoding models. 
MindAdapter adopts a decoupled linear-residual cascade alignment paradigm by freezing a pretrained explicit brain functional alignment backbone (coarse) and introducing a lightweight nonlinear residual adapter (fine), thereby disentangling global cross-subject correspondence from subject-specific residual corrections for fine-grained spatial and semantic calibration. To further preserve global representational stability, we design a topology-anchored dual-stream manifold constraint, where a small set of shared stimuli serves as \emph{topological pins} with voxel-level paired supervision, while a semantic stream enforces consistency through a frozen vision–language decoder on unpaired brain data. Together, MindAdapter efficiently injects subject-specific corrections while maintaining the global representational geometry learned during pretraining. Experiments on the Natural Scenes Dataset (NSD) demonstrate that MindAdapter substantially improves cross-subject visual reconstruction and retrieval accuracy using only a few shared stimuli, offering a practical and data-efficient solution for personalized brain-to-visual decoding.

\end{abstract}


\maketitle

\section{Introduction}
% Decoding visual experiences from human brain activity has become a central goal of computational neuroscience and machine learning, with recent progress enabling the reconstruction and retrieval of perceived images from fMRI signals 
Decoding visual experiences from human brain activity is a central goal in computational neuroscience and machine learning, with recent advances enabling image reconstruction and retrieval from fMRI signals.~\cite{Naselaris2011,daimindaligner,Qian2020,Horikawa2017,bhattacharjee2025aligning}.
Beyond single-subject decoding, a more challenging and practically relevant problem is \emph{cross-subject} brain-to-visual decoding, where a model trained on one individual must generalize to another with different neural representations, voxel layouts, and response statistics \cite{mindeyev1}.
This setting is crucial for real-world deployment, as collecting large-scale, image-aligned fMRI for every new subject is costly, time-consuming, and often infeasible \cite{zhou2025rest2visual}.
However, inter-subject variability in cortical organization and neural tuning induces severe distribution shift, causing models that perform well within a subject to degrade substantially when transferred across subjects \cite{daimindaligner}.
Bridging this representational gap between brains therefore remains a fundamental obstacle for scalable visual brain decoding \cite{mindeyev2}.

% Decoding visual experiences from human brain activity has recently achieved remarkable progress, enabling the reconstruction and retrieval of perceived images from fMRI signals~\cite{Naselaris2011,daimindaligner,Qian2020,Horikawa2017}.
% Beyond single-subject settings, a more challenging and practically important problem is \emph{cross-subject} brain-to-visual decoding, where models must generalize across individuals with different neural representations and voxel organizations~\cite{mindeyev1}.
% This setting is essential for real-world deployment, as collecting large-scale, image-aligned fMRI data for each new subject is costly and often infeasible~\cite{zhou2025rest2visual}.
% However, substantial inter-subject variability induces severe distribution shift, causing a dramatic performance drop when models are transferred across subjects.
% Bridging this representational gap therefore remains a fundamental challenge for scalable visual brain decoding~\cite{daimindaligner,mindeyev2}.

\begin{figure}[t!]
\centering
\includegraphics[width=\columnwidth]{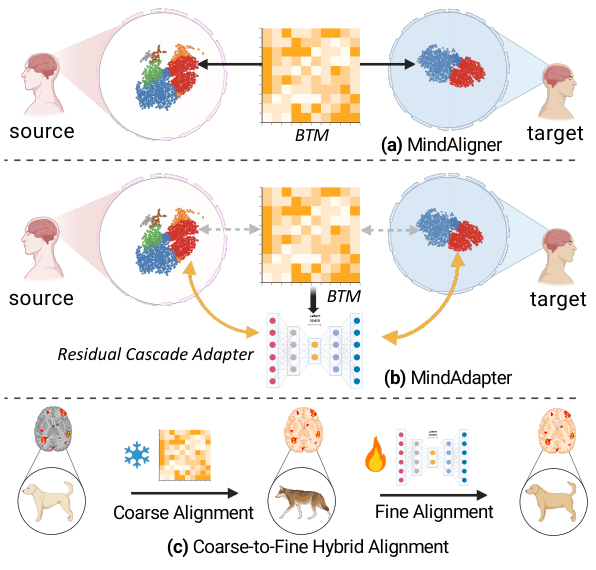}
\vspace{-2em}
\caption{
(a) MindAligner performs cross-subject alignment using a linear BTM.
(b) MindAdapter augments the frozen BTM with a lightweight nonlinear residual adapter for few-shot calibration.
(c) Coarse-to-fine hybrid alignment: linear coarse mapping followed by nonlinear refinement.
} 
\label{fig:coarsetofine}
\vspace{-1.5em}
\end{figure}

To address cross-subject variability, recent approaches have shifted from direct image reconstruction to representation-level alignment, mapping brain activity from different individuals into a shared latent or visual-semantic space \cite{seeliger2018generative, Minddiffuser, Ozcelik2023,eccv25}.
Representative frameworks such as MindAligner \cite{daimindaligner}, MindBridge \cite{mindbridge}, and related cross-subject decoders learn a subject-conditional brain transformation that aligns fMRI responses to a common visual embedding, typically defined by pretrained vision or vision--language models \cite{daimindaligner,mindbridge,unibrain,tian2025brainguard}.
This paradigm enables zero-shot transfer across subjects without requiring them to observe the same stimuli during training, and has led to substantial improvements in cross-subject retrieval and reconstruction \cite{mindeyev2,willsaligner}.
However, these models rely on a fixed pretrained alignment that must accommodate large inter-subject differences using only population-level statistics.
As a result, even after large-scale training, residual subject-specific mismatches persist, limiting decoding fidelity and robustness when applied to new individuals ({\autoref{fig:coarsetofine}}).

% 2. 范式进化：从“矩阵学习”到“基础模型适配（PEFT）”
% • MindAligner的做法： 它需要为每一个新受试者重新训练一个大脑转移矩阵（BTM），尽管它是低秩分解的，但本质上仍是从头学习映射关系。
% • 将MindAligner视为一个**“大脑功能底座模型”，通过冻结主体权重并引入Adapter，证明了显式功能对齐特征的通用性（Generalizability）**。
% • 提出首个针对“显式脑功能对齐”的参数高效微调（PEFT）方案，定义了脑解码领域的“基础模型+轻量适配”新范式。

A key limitation of current cross-subject decoding frameworks is that they treat alignment as a one-time, globally optimized mapping, implicitly assuming that the pretrained transformation generalizes equally well to all individuals \cite{daimindaligner}.
In practice, however, fine-grained functional topographies and idiosyncratic neural representations vary substantially across subjects, leading to systematic but low-dimensional residual misalignment even after population-level training \cite{mindtuner} (\autoref{fig:coarsetofine}). 
Importantly, such mismatches are not random: they manifest as consistent distortions in the predicted visual embedding space, which degrade both retrieval accuracy and image reconstruction quality.
This suggests that cross-subject decoding is not fundamentally limited by the lack of data, but rather by the absence of a mechanism for subject-specific calibration.
A small number of shared stimuli between two subjects provides precisely the information needed to estimate this residual alignment error, offering a principled opportunity to refine pretrained mappings without retraining large-scale models \cite{mindtuner}.

% Motivated by this observation, we propose \textit{MindAdapter}, a few-shot calibration framework for pretrained cross-subject brain-to-visual decoding models.
% Instead of retraining or modifying the original cross-subject mapper, MindAdapter freezes the pretrained alignment network and introduces a lightweight nonlinear adapter that estimates and corrects the subject-specific residual misalignment using only a handful of shared stimuli.
% The adapter is optimized with a hybrid objective that anchors voxel-level predictions on paired few-shot data while preserving semantic consistency on large-scale unpaired data via CLIP-based visual embeddings, enabling stable adaptation without overfitting.
% This design allows MindAdapter to inherit the global structure learned by population-level models while injecting individualized corrections that significantly improve decoding fidelity. Our main contributions are threefold:

Motivated by these observations, we propose \textit{MindAdapter}, a parameter-efficient few-shot calibration framework for pretrained cross-subject brain-to-visual decoding models. 
Unlike existing approaches that rely on time-intensive subject-specific fine-tuning (e.g., MindEye2’s one-hour protocol), MindAdapter introduces a \emph{Decoupled Linear--Residual Cascade} paradigm. 
Specifically, we freeze a pretrained explicit alignment backbone based on the Brain Transfer Matrix (BTM) \cite{daimindaligner} that captures global cross-subject functional mappings, and augment it with a lightweight nonlinear residual adapter that estimates and corrects subject-specific misalignment. 
This design mathematically disentangles universal brain manifolds from individual residual deviations, enabling high-fidelity adaptation with minimal data while mitigating catastrophic forgetting.
To ensure stable adaptation under extreme few-shot supervision, MindAdapter is optimized with a \emph{Topology-Anchored Dual-Stream} constraint (i.e., anchor-guided geometric stabilization under sparse supervision). 
A small set of paired shared stimuli is treated as \emph{topological pins} that lock the geometric structure of the novel subject’s fMRI manifold through an \emph{Anchor Stream} with voxel-level losses. 
Meanwhile, a \emph{Semantic Stream} enforces global representational consistency on unpaired data via frozen CLIP-based visual embeddings, preventing semantic drift in data-scarce regimes. 
Together, MindAdapter inherits the population-level representational structure learned during pretraining while injecting individualized residual corrections, leading to substantially improved cross-subject visual decoding fidelity. Our main contributions are threefold:

\begin{itemize}
    % \item We identify subject-specific residual misalignment as a key bottleneck in pretrained cross-subject brain-to-visual decoding models, and show that it can be efficiently corrected using only a few shared stimuli.
    % \item We propose \textit{MindAdapter}, a lightweight few-shot calibration module that refines frozen cross-subject decoders via a nonlinear residual adapter without retraining large models.
    % \item We develop a hybrid optimization scheme that combines paired voxel anchors with unpaired CLIP-based semantic regularization, enabling robust and data-efficient cross-subject adaptation.
% \item We reveal that \emph{subject-specific residual misalignment} constitutes a fundamental bottleneck in pretrained cross-subject brain-to-visual decoding, and demonstrate that such residuals can be effectively corrected with only a handful of shared stimuli.

% \item We propose \textit{MindAdapter}, a parameter-efficient few-shot calibration framework that instantiates a \emph{Decoupled Linear--Residual Cascade} by augmenting a frozen explicit alignment backbone with a lightweight nonlinear residual adapter, enabling fine-grained subject-specific calibration without retraining large models.

\item We propose \textit{MindAdapter}, a parameter-efficient few-shot calibration framework that instantiates a \emph{coarse-to-fine Decoupled Linear--Residual Cascade} by augmenting a frozen explicit alignment backbone (coarse) with a lightweight nonlinear residual adapter (fine), enabling fine-grained subject-specific calibration without retraining large models.

\item We introduce a \emph{Topology-Anchored Dual-Stream} optimization scheme that jointly enforces voxel-level anchor supervision on paired shared samples and CLIP-based semantic consistency on unpaired data, yielding robust and data-efficient cross-subject adaptation.

\item Extensive experiments on the NSD demonstrate that MindAdapter substantially improves cross-subject visual reconstruction and retrieval accuracy under extremely limited shared stimuli, achieving strong performance across multiple subject pairs and evaluation metrics.
\end{itemize}

\begin{figure*}[t!]
\centering
\includegraphics[width=0.8\textwidth]{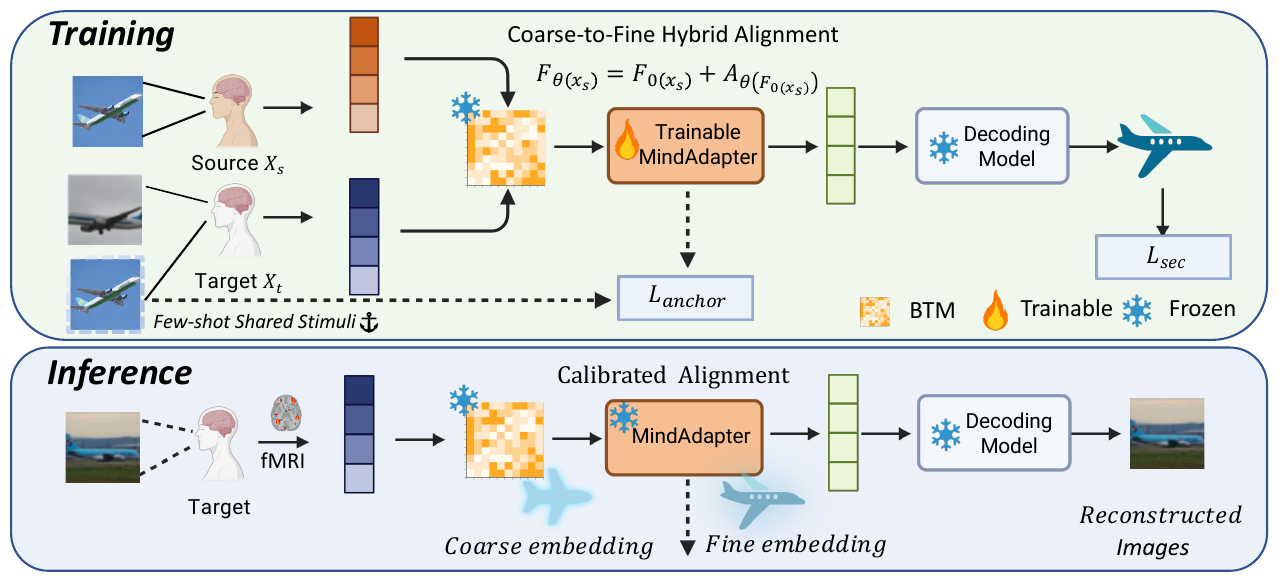}
\vspace{-1em}
\caption{
Overview of MindAdapter.
During training, a frozen pretrained BTM provides a coarse cross-subject alignment, while a lightweight nonlinear MindAdapter learns residual corrections using a small set of shared stimuli through anchor-based supervision and a frozen visual decoder. 
% The resulting mapping follows a coarse-to-fine formulation $F_\theta(x_s)=F_0(x_s)+A_\theta(F_0(x_s))$. 
During inference, target-subject fMRI is mapped to calibrated latent embeddings and decoded into images using the frozen decoder, enabling high-fidelity cross-subject visual reconstruction.
} 
\label{fig:pipeline}
\vspace{-1.5em}
\end{figure*}

\section{Related Work}
\subsection{Cross-Subject Brain Decoding}
A central challenge in brain decoding is the strong inter-subject variability of neural representations, which severely limits the generalization of subject-specific decoders \cite{shenneuro,mentzelopoulos2024neural}.
Early work addressed this problem through anatomical normalization and voxel-wise linear mappings across subjects, but these approaches struggle to capture higher-order functional correspondence ~\cite{Horikawa2017}.
More recent studies have proposed learning shared latent spaces across subjects using deep neural networks, canonical correlation analysis, or hyperalignment-style transformations, enabling cross-subject transfer of encoding and decoding models \cite{difumo, rastegarnia2023brain}.
Building on large-scale datasets such as NSD ~\cite{nsd}, several recent methods further integrate deep visual backbones and pretrained multimodal models (e.g., CLIP) \cite{Radford2021} to learn population-level brain-to-visual mappings that can generalize across individuals \cite{Lin2022, Takagi2023, maiunibrain, mindeyev1, chen2023seeing}.
Specifically, MindEye2~\cite{mindeyev2} aligns brain signals from multiple subjects into a shared latent space via ridge regression, followed by a common decoding module.
MindBridge~\cite{mindbridge} constructs pseudo stimuli to create shared stimulus pairs for cross-subject alignment.
MindAligner~\cite{daimindaligner} performs soft functional alignment across subjects by learning a Brain Transfer Matrix that enables the transfer of pretrained decoders to novel subjects.
Collectively, these methods mitigate inter-subject variability through anatomical normalization or coarse-grained linear functional alignment.
While they achieve strong zero-shot cross-subject decoding performance, they implicitly assume that a single global alignment is sufficient to account for individual differences.
As a result, higher-order and subject-specific residual misalignment remains unmodeled, limiting the fidelity of cross-subject visual decoding~\cite{mindeyev1,mindeyev2,mindbridge}.

\subsection{Few-shot Calibration for Pretrained Models}
Beyond training universal population-level models, a growing body of work has explored how to adapt pretrained networks to new domains or individuals using limited data \cite{liu2023parameter,chen2023sam}.
In computer vision and vision--language models, few-shot adaptation and test-time training techniques such as adapters \cite{rebuffi2017learning}, LoRA \cite{hu2022lora}, prompt tuning \cite{lester2021power}, and entropy or consistency regularization have been shown to substantially improve robustness and personalization without retraining the entire model \cite{sung2022vl,liu2025kpl,mao2025survey,mindtuner}.
In neuroscience, MindTuner~\cite{mindtuner} models subject-specific \emph{visual fingerprints} via LoRA/Skip-LoRA and employs an image-mediated pivot module for semantic calibration from fMRI to text.
However, existing few-shot brain adaptation methods typically operate directly on the full decoder or mapping network, which risks overfitting and breaking the pretrained cross-subject geometry \cite{mindshot,mindeyev2}.
In contrast, our work formulates few-shot adaptation as a lightweight residual calibration problem on top of a frozen cross-subject brain-to-visual decoder, allowing subject-specific corrections to be learned while preserving the global population-level structure.

\begin{figure}[t!]
\centering
\includegraphics[width=0.8\columnwidth]{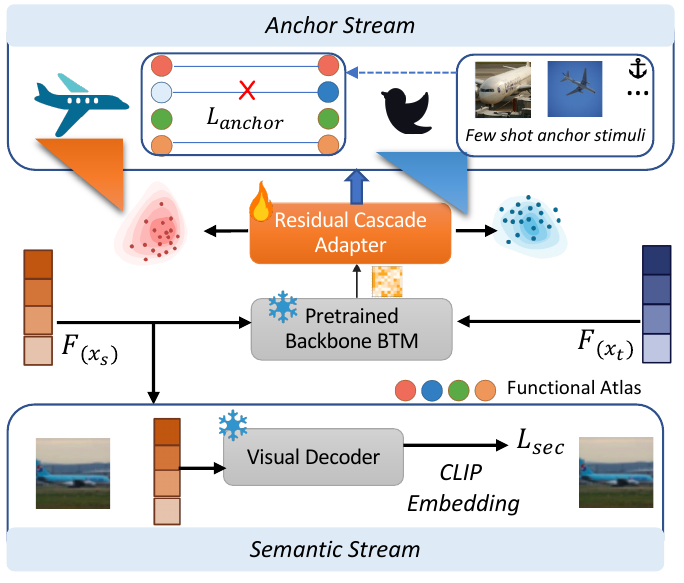}
\vspace{-1em}
\caption{
Topology-anchored dual-stream manifold constraint.
MindAdapter jointly optimizes an anchor-based alignment loss $L_{\text{anchor}}$ on few-shot paired samples and a semantic consistency loss $L_{\text{sec}}$ on unpaired data. 
This dual-stream design stabilizes the pretrained functional manifold while enabling subject-specific residual adaptation.
} 
\label{fig:dualstream}
\vspace{-1em}
\end{figure}

\section{Methodology}

\subsection{Problem Setup and Notation}

We study cross-subject brain-to-visual decoding, where the goal is to transfer neural representations between different human subjects in order to enable visual reconstruction or retrieval across individuals.
Let $s$ denote a \emph{source} subject and $t$ denote a \emph{target} subject.
Each subject observes visual stimuli $I \in \mathcal{I}$ while their corresponding brain responses are recorded as fMRI voxel patterns.

The fMRI measurements of subject $s$ and $t$ are denoted as
\begin{equation}
    \mathbf{x}_s \in \mathbb{R}^{d_s}, \qquad \mathbf{x}_t \in \mathbb{R}^{d_t},
\end{equation}
where $d_s$ and $d_t$ are the voxel dimensions of the two subjects.
Each stimulus $I$ is associated with a visual semantic representation extracted by a pretrained vision--language model (e.g., CLIP),
\begin{equation}
    \mathbf{c}(I) \in \mathbb{R}^{d_c}.
\end{equation}

We assume access to \emph{unpaired} brain--image datasets for each subject,
\begin{equation}
    \mathcal{D}^{s}_{\mathrm{train}} = \{ (\mathbf{x}_s^{i}, I^{i}) \}_{i=1}^{N_s}, 
    \qquad
    \mathcal{D}^{t}_{\mathrm{train}} = \{ (\mathbf{x}_t^{j}, J^{j}) \}_{j=1}^{N_t},
\end{equation}
where the stimulus sets $\{I^{i}\}$ and $\{J^{j}\}$ are disjoint and the two subjects never observe the same images during pretraining.

In addition, a small \emph{few-shot} set of shared stimuli is available,
\begin{equation}
    \mathcal{D}_{\mathrm{fs}} = \{ (\mathbf{x}_s^{k}, \mathbf{x}_t^{k}, I^{k}) \}_{k=1}^{M}, \quad M \ll N_s, N_t,
\end{equation}
where both subjects viewed the same image $I^k$.
This few-shot set is used only for calibration and does not overlap with the final evaluation images.

At test time, we are given unseen brain responses from the source subject,
\begin{equation}
    \mathcal{D}_{\mathrm{test}} = \{ (\mathbf{x}_s^{\ell}, I^{\ell}) \}_{\ell=1}^{N_{\mathrm{test}}},
\end{equation}
and the task is to decode their visual content without access to $\mathbf{x}_t$.

The goal is to learn a cross-subject brain mapping and a visual decoding function such that
\begin{equation}
    \mathbf{x}_s \;\xrightarrow{\;\;F\;\;} \; \hat{\mathbf{x}}_t 
    \;\xrightarrow{\;\;G\;\;} \; \hat{\mathbf{c}}
    \;\approx\; \mathbf{c}(I),
\end{equation}
where $F$ transfers brain activity from subject $s$ to subject $t$, and $G$ projects target-subject brain responses into the visual semantic space.
Few-shot data $\mathcal{D}_{\mathrm{fs}}$ is used to calibrate $F$ without accessing $\mathcal{D}_{\mathrm{test}}$.

\begin{algorithm}[t]
\caption{MindAdapter}
\label{alg:mindadapter}
\begin{algorithmic}[1]

\STATE \textbf{Input:} Pretrained cross-subject mapper $F_0:\mathbf{x}_s \rightarrow \hat{\mathbf{x}}_t$, 
frozen visual decoder $G$, 
few-shot paired set $\mathcal{P}=\{(\mathbf{x}_s^{(i)},\mathbf{x}_t^{(i)})\}_{i=1}^m$, 
unpaired source set $\mathcal{U}=\{(\mathbf{x}_s^{(j)},\mathbf{c}_s^{(j)})\}_{j=1}^N$, 
learning rate $\eta$, loss weights $\alpha,\lambda$.

\STATE Initialize residual adapter $A_\theta$ and define
$F_\theta(\mathbf{x}_s)=F_0(\mathbf{x}_s)+A_\theta(F_0(\mathbf{x}_s))$.
Freeze parameters of $F_0$.
\FOR{each training iteration}
    \STATE Sample a minibatch $\mathcal{B}_p \subset \mathcal{P}$ and $\mathcal{B}_u \subset \mathcal{U}$.

    \STATE \textbf{Paired Anchor Stream:}
    \FOR{each $(\mathbf{x}_s^{(i)},\mathbf{x}_t^{(i)}) \in \mathcal{B}_p$}
        \STATE $\hat{\mathbf{x}}_t^{(i)} \leftarrow F_\theta(\mathbf{x}_s^{(i)})$
        \STATE Compute
        $
        % \ell_{\text{anchor}}^{(i)}
        % =
        % \|\hat{\mathbf{x}}_t^{(i)}-\mathbf{x}_t^{(i)}\|^2
        % +
        % \alpha\,\ell_{\text{NCE}}\!\big(\hat{\mathbf{x}}_t^{(i)},\mathbf{x}_t^{(i)}\big)
        \mathcal{L}_{\text{anchor}}
=
\sum_{i=1}^m \alpha
\|F_\theta(\mathbf{x}_s^{(i)})-\mathbf{x}_t^{(i)}\|^2
+
 \,\ell_{\text{NCE}}\!\big(F_\theta(\mathbf{x}_s^{(i)}),\mathbf{x}_t^{(i)}\big).
        $
    \ENDFOR
    \STATE $\mathcal{L}_{\text{anchor}} \leftarrow \frac{1}{|\mathcal{B}_p|}\sum_{i\in \mathcal{B}_p}\ell_{\text{anchor}}^{(i)}$

    \STATE \textbf{Unpaired Semantic Stream:}
    \FOR{each $(\mathbf{x}_s^{(j)},\mathbf{c}_s^{(j)}) \in \mathcal{B}_u$}
        \STATE $\hat{\mathbf{x}}_t^{(j)} \leftarrow F_\theta(\mathbf{x}_s^{(j)})$
        \STATE $\hat{\mathbf{c}}^{(j)} \leftarrow G(\hat{\mathbf{x}}_t^{(j)})$
        \STATE Compute
        $
        % \ell_{\text{semantic}}^{(j)}
        % =
        % \|\hat{\mathbf{c}}^{(j)}-\mathbf{c}_s^{(j)}\|^2
        \ell_{\text{sec}}^{(j)} =
\ell_{\text{NCE}}\!\big(\hat{\mathbf{c}}^{(j)}, \mathbf{c}_s^{(j)}\big)
        $
        
    \ENDFOR
    \STATE $\mathcal{L}_{\text{sec}} \leftarrow \frac{1}{|\mathcal{B}_u|}\sum_{j\in \mathcal{B}_u}\ell_{\text{sec}}^{(j)}$

    \STATE \textbf{Joint Update:}
    \STATE $\mathcal{L} \leftarrow \lambda\mathcal{L}_{\text{anchor}} +  \mathcal{L}_{\text{sec}}$
    \STATE $\theta \leftarrow \theta - \eta \nabla_\theta \mathcal{L}$
\ENDFOR

\STATE \textbf{Output:} Calibrated mapper $F_\theta = F_0 + A_\theta$.

\end{algorithmic}
\end{algorithm}
% \vspace{-1em}

\subsection{Linear Cascade Backbone}

Let a pretrained cross-subject brain alignment model be
\begin{equation}
F_0:\mathbb{R}^{d_s}\rightarrow\mathbb{R}^{d_t},
\end{equation}
mapping fMRI from a source subject to a target subject, and let a frozen visual decoder
\begin{equation}
G:\mathbb{R}^{d_t}\rightarrow\mathbb{R}^{d_c}
\end{equation}
map target-subject brain activity into a visual-semantic embedding space.
The pretrained decoding pipeline is therefore
\begin{equation}
\hat{\mathbf{c}} = G(F_0(\mathbf{x}_s)).
\end{equation}

In existing cross-subject decoders, $F_0$ is learned by minimizing voxel-space error,
\begin{equation}
\min_F \; \mathbb{E}\big[\|F(\mathbf{x}_s)-\mathbf{x}_t\|^2\big],
\end{equation}
without directly optimizing performance in the downstream visual space.
This creates a fundamental mismatch: the metric used to train $F_0$ is Euclidean in voxel space, while the task objective is semantic similarity after a nonlinear mapping $G$.

This mismatch is not benign.
Let the true target be $\mathbf{x}_t$ and define the voxel residual
\begin{equation}
\Delta = F_0(\mathbf{x}_s)-\mathbf{x}_t.
\end{equation}
After decoding, the visual error becomes
\begin{equation}
\|G(F_0(\mathbf{x}_s))-G(\mathbf{x}_t)\|
\approx \|J_G(\mathbf{x}_t)\Delta\|,
\end{equation}
where $J_G$ is the Jacobian of $G$.
Since $J_G$ is highly anisotropic, voxel errors that are small in $\ell_2$ norm can be greatly amplified along visually sensitive directions.
Therefore, minimizing $\|\Delta\|^2$ does not guarantee small visual-semantic error.

Moreover, cross-subject brain representations are not related by a linear map.
Let $F^*$ denote the true subject-to-subject mapping; then the pretrained model satisfies
\begin{equation}
F_0(\mathbf{x}_s)=F^*(\mathbf{x}_s)-\mathbf{r}(\mathbf{x}_s),
\end{equation}
where $\mathbf{r}(\mathbf{x}_s)$ is a structured, stimulus-dependent residual.
Because $F_0$ is trained on population-level data, this residual is systematic rather than random, and it persists even with large training sets.

As a result, pretrained cross-subject decoders are inherently misaligned with the frozen visual decoder $G$, leading to biased visual predictions that cannot be eliminated by further voxel-level training.
% This limitation motivates a calibration mechanism that directly corrects the semantic error induced by $F_0$ under $G$, rather than attempting to relearn the entire cross-subject mapping.
This limitation motivates a coarse-to-fine calibration mechanism that directly corrects the semantic error induced by $F_0$ under $G$ via a lightweight residual pathway, rather than attempting to relearn the entire cross-subject mapping (\autoref{fig:pipeline}).

\begin{figure*}[ht!]
\centering
\includegraphics[width=0.75\textwidth]{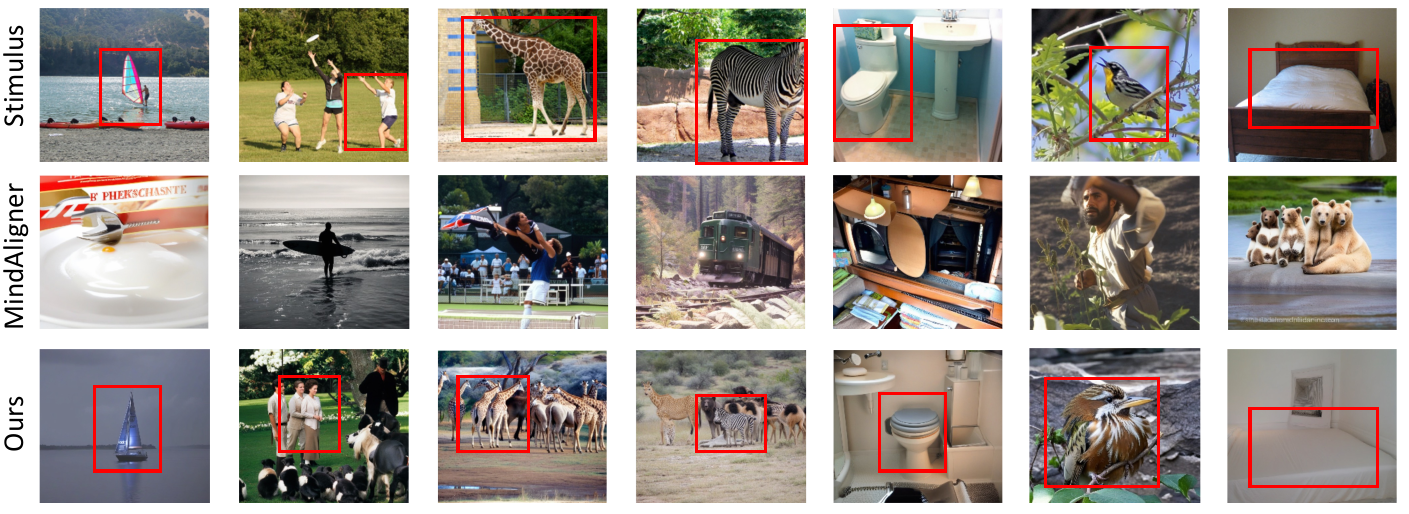}
\vspace{-1em}
\caption{
Qualitative comparison of reconstructed images. MindAdapter produces reconstructions with improved object localization and semantic fidelity over MindAligner, demonstrating the effectiveness of few-shot residual calibration.
} 
\label{vis1}
\vspace{-1.5em}
\end{figure*}

\subsection{Residual Cascade Adapter}

Let $F_0$ be the pretrained cross-subject brain mapper defined in the previous section.
We introduce a lightweight residual adapter
\begin{equation}
A_\theta:\mathbb{R}^{d_t}\rightarrow\mathbb{R}^{d_t},
\end{equation}
and define the calibrated mapping as
\begin{equation}
F_\theta(\mathbf{x}_s)=F_0(\mathbf{x}_s)+A_\theta(F_0(\mathbf{x}_s)).
\end{equation}
Rather than relearning the full cross-subject transform, $A_\theta$ learns a small correction in the target-subject voxel space.

From the analysis above, the pretrained model produces a structured residual
\begin{equation}
F_0(\mathbf{x}_s)=F^*(\mathbf{x}_s)-\mathbf{r}(\mathbf{x}_s),
\end{equation}
where $\mathbf{r}$ induces large errors after the frozen decoder $G$.
The goal of MindAdapter is therefore to learn
\begin{equation}
A_\theta(F_0(\mathbf{x}_s))\approx \mathbf{r}(\mathbf{x}_s),
\end{equation}
so that $F_\theta(\mathbf{x}_s)\approx F^*(\mathbf{x}_s)$ in the semantic directions that matter for $G$.
% Given a few shared stimuli $\{(\mathbf{x}_s^{(i)},\mathbf{x}_t^{(i)},\mathbf{c}^{(i)})\}_{i=1}^m$, where $\mathbf{c}^{(i)}=G(\mathbf{x}_t^{(i)})$
% Given a few shared stimuli 
% $\{(\mathbf{x}_s^{(i)},\mathbf{x}_t^{(i)},\mathbf{c}^{(i)})\}_{i=1}^m$,
% where $\mathbf{c}^{(i)} = E_{\mathrm{CLIP}}(I^{(i)})$ is extracted by a frozen CLIP image encoder. We optimize
% \begin{equation}
% \min_\theta \sum_{i=1}^m \alpha\|F_\theta(\mathbf{x}_s^{(i)})-\mathbf{x}_t^{(i)}\|^2
% + \sum_{i=1}^m 
% \ell_{\text{NCE}}\big(F_\theta(\mathbf{x}_s^{(i)}), \mathbf{x}_t^{(i)}\big) .
% \end{equation}
% The first term anchors the adapter to the true target-subject voxels, while the second term enforces semantic consistency under the frozen decoder. $\ell_{\text{NCE}}$ denotes an InfoNCE-style contrastive loss.
Given a few shared stimuli 
$\{(\mathbf{x}_s^{(i)},\mathbf{x}_t^{(i)})\}_{i=1}^m$,
where both subjects observe the same images, we optimize the residual adapter using a voxel-level anchor objective:
\begin{equation}
\min_\theta 
\sum_{i=1}^m 
\alpha\,\|F_\theta(\mathbf{x}_s^{(i)})-\mathbf{x}_t^{(i)}\|^2
\;+\;
\sum_{i=1}^m 
\ell_{\text{NCE}}\!\big(F_\theta(\mathbf{x}_s^{(i)}), \mathbf{x}_t^{(i)}\big).
\end{equation}
The first term enforces direct voxel-wise anchoring to the ground-truth target-subject responses, providing a strong geometric constraint.
The second term applies an InfoNCE-style contrastive loss in the target-subject voxel space, encouraging discriminative alignment between predicted and true voxel representations.
Together, these paired anchors act as sparse but reliable topological constraints that stabilize residual calibration under extremely limited supervision.

Importantly, because $A_\theta$ is applied after $F_0$, the optimization operates on the residual subspace.
Under a first-order approximation,
\begin{equation}
G(F_\theta(\mathbf{x}_s))\approx G(F_0(\mathbf{x}_s)) + J_G(F_0(\mathbf{x}_s))A_\theta(F_0(\mathbf{x}_s)),
\end{equation}
so the adapter directly corrects the visually amplified error directions identified by $J_G$.
This allows a small number of paired samples to efficiently suppress the dominant semantic bias, even when $F_0$ remains frozen.
Therefore, MindAdapter transforms few-shot shared stimuli into a targeted semantic calibration of a large pretrained cross-subject brain decoder, achieving visual alignment without retraining the full mapping (\autoref{fig:pipeline}).

\subsection{Topology-Anchored Dual-Stream Loss}

The few-shot setting presents a fundamental optimization challenge.
The paired set $\mathcal{P}=\{(\mathbf{x}_s^{(i)},\mathbf{x}_t^{(i)},\mathbf{c}^{(i)})\}_{i=1}^m$ is extremely small and prone to overfitting, while the unpaired source data $\mathcal{U}=\{(\mathbf{x}_s^{(j)},\mathbf{c}_s^{(j)})\}_{j=1}^N$ is abundant but lacks target-subject voxel supervision.
Training only on $\mathcal{P}$ would fit the few anchors but distort the global semantic geometry, whereas training only on $\mathcal{U}$ cannot resolve the cross-subject shift.
MindAdapter resolves this tension by jointly optimizing two complementary objective streams (\autoref{fig:dualstream}).

% \paragraph{Paired Anchor Stream.}
% For each shared stimulus, we enforce voxel-level and retrieval-level consistency
% \begin{equation}
% \mathcal{L}_{\text{anchor}}
% =
% \sum_{i=1}^m 
% \|F_\theta(\mathbf{x}_s^{(i)})-\mathbf{x}_t^{(i)}\|^2
% +
% \alpha \,\ell_{\text{NCE}}\!\big(F_\theta(\mathbf{x}_s^{(i)}),\mathbf{x}_t^{(i)}\big),
% \end{equation}
% where $\ell_{\text{NCE}}$ is a contrastive loss in the target-subject voxel space.
% This term ensures that the residual adapter learns the correct cross-subject correspondence for the few available ground-truth pairs.
% These paired samples act as sparse \emph{topological pins} that anchor the geometry of the novel subject’s voxel manifold.
\paragraph{Paired Anchor Stream.}
For each shared stimulus, we enforce voxel-level and retrieval-level consistency
\begin{equation}
\mathcal{L}_{\text{anchor}}
=
\sum_{i=1}^m 
\alpha\,
\underbrace{\|F_\theta(\mathbf{x}_s^{(i)})-\mathbf{x}_t^{(i)}\|^2}_{\boldsymbol{\mathcal{L}}_{\text{Anchor}}^{\text{MSE}}}
+
\underbrace{\ell_{\text{NCE}}\!\big(F_\theta(\mathbf{x}_s^{(i)}),\mathbf{x}_t^{(i)}\big)}_{\boldsymbol{\mathcal{L}}_{\text{Anchor}}^{\text{NCE}}},
\end{equation}
% \begin{equation}
% \mathcal{L}_{\text{anchor}}
% =
% \sum_{i=1}^m 
% \alpha\,
% \underbrace{\|F_\theta(\mathbf{x}_s^{(i)})-\mathbf{x}_t^{(i)}\|^2}_{\boldsymbol{\mathcal{L}}_{\text{Anchor}}^{\text{MSE}}}
% +
% \underbrace{\ell_{\text{NCE}}\!\big(F_\theta(\mathbf{x}_s^{(i)}),\mathbf{x}_t^{(i)}\big)}_{\boldsymbol{\mathcal{L}}_{\text{Anchor}}^{\text{NCE}}}
% \end{equation}
where $\ell_{\text{NCE}}$ is a contrastive loss in the target-subject voxel space.
This term ensures that the residual adapter learns the correct cross-subject correspondence for the few available ground-truth pairs.
These paired samples act as sparse \emph{topological pins} that anchor the geometry of the novel subject’s voxel manifold.

\begin{table*}[t]
\centering
\caption{
Comparison between MindAligner and MindAdapter. $\uparrow$ indicates higher is better and $\downarrow$ indicates lower is better.
}
\vspace{-1em}
\label{tab:64shot_comparison}
\resizebox{0.75\textwidth}{!}{
\begin{tabular}{@{}lcccccccc@{}}
\toprule
\multirow{2}{*}{\textbf{Method}} 
& \multicolumn{4}{c}{\textbf{Low-Level Metrics}} 
& \multicolumn{4}{c}{\textbf{High-Level Metrics}} \\
\cmidrule(lr){2-5} \cmidrule(lr){6-9}
& PixCorr$\uparrow$ 
& SSIM$\uparrow$ 
& AlexNet(2)$\uparrow$ 
& AlexNet(5)$\uparrow$ 
& Inception$\uparrow$ 
& CLIP$\uparrow$ 
& EffNet-B$\downarrow$ 
& SwAV$\downarrow$ \\
\midrule
MindAligner (Subj.1)      
& .2136 & .4136 & 87.64\% & 92.55\% & 83.19\% & 81.60\% & .8004 & .459 \\
\rowcolor{gray!12}
\textbf{Ours (Subj.1)}    
& \textbf{.2239} & \textbf{.4188} & \textbf{88.43\%} & \textbf{93.05\%} & \textbf{84.30\%} & \textbf{81.83\%} & \textbf{.7896} & \textbf{.451} \\
\midrule
MindAligner (Subj.2)      
& .2196 & \textbf{.4274} & 88.39\% & 93.93\% & 83.94\% & 83.12\% & .7900 & .449 \\
\rowcolor{gray!12}
\textbf{Ours (Subj.2)}    
& \textbf{.2198} & .4245 & \textbf{88.66\%} & \textbf{94.14\%} & \textbf{84.12\%} & \textbf{83.77\%} & \textbf{.7861} & \textbf{.445} \\
\midrule
MindAligner (Subj.5)      
& .1936 & .4083 & \textbf{85.11\%} & 91.59\% & \textbf{84.69\%} & 83.27\% & \textbf{.7776} & \textbf{.449} \\
\rowcolor{gray!12}
\textbf{Ours (Subj.5)}    
& \textbf{.1996} & \textbf{.4123} & 85.04\% & \textbf{92.06\%} & 84.10\% & \textbf{83.63\%} & .7817 & .450 \\
\midrule
MindAligner (Subj.7)      
& .1702 & .4069 & 81.49\% & 88.09\% & 79.46\% & 76.71\% & .8356 & {.486} \\
\rowcolor{gray!12}
\textbf{Ours (Subj.7)}    
& \textbf{.1784} & \textbf{.4107} & \textbf{82.19\%} & \textbf{88.67\%} & \textbf{79.68\%} & \textbf{77.28\%} & \textbf{.8290} & \textbf{.483} \\
\bottomrule
\end{tabular}}
\vspace{-1em}
\end{table*}

\paragraph{Unpaired Semantic Stream.}
For each unpaired source voxel $\mathbf{x}_s^{(j)}$, we enforce semantic consistency in the visual space
\begin{equation}
\mathcal{L}_{\text{sec}}
=
\sum_{j=1}^N 
\ell_{\text{NCE}}\!\big(G(F_\theta(\mathbf{x}_s^{(j)})),\mathbf{c}_s^{(j)}\big)
\end{equation}
where $\mathbf{c}_s^{(j)}$ is the CLIP embedding of the image seen by the source subject.
This term preserves the pretrained model’s global visual geometry and prevents degenerate solutions that satisfy the few-shot anchors but collapse semantic structure.

\paragraph{Joint Optimization.}
The final objective is
\begin{equation}
\min_\theta 
\mathcal{L}(\theta)
=
\lambda \mathcal{L}_{\text{anchor}}
+
 \mathcal{L}_{\text{sec}} .
\end{equation}

This two-stream design has a clear geometric interpretation.
The anchor stream constrains $F_\theta$ to pass through the true target-subject voxels on the shared stimuli, while the semantic stream constrains $F_\theta$ to remain on the pretrained visual manifold induced by $G$.
Under the first-order analysis in the previous section, the semantic loss projects the learned residual $A_\theta$ onto the dominant visual-error subspace, while the anchor loss fixes its scale and direction.
As a result, even with very few paired samples, the adapter converges to the unique residual that corrects the cross-subject semantic bias without destroying the pretrained alignment.

\begin{figure*}[t!]
\centering
\includegraphics[width=\textwidth]{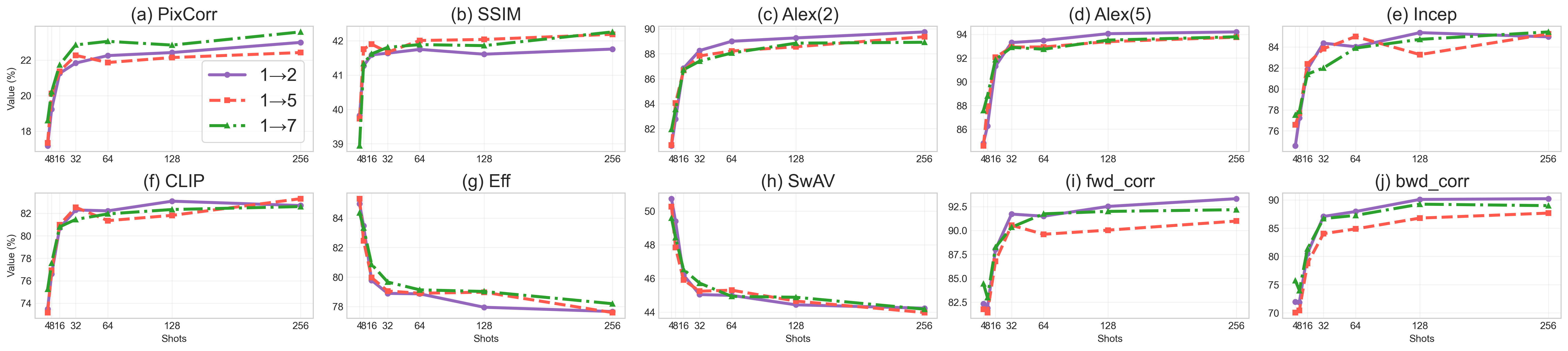}
\vspace{-2em}
\caption{
Performance scaling of MindAdapter as a function of the number of few-shot anchor samples for different cross-subject transfer directions (1→2, 1→5, 1→7), evaluated by low-level image fidelity, high-level semantic similarity, and retrieval accuracy. {fwd\_corr}: brain-to-image retrieval accuracy; 
{bwd\_corr}: image-to-brain retrieval accuracy.
} 
\label{fig:fewshotmetric_1}
\vspace{-1.5em}
\end{figure*}

% 神经科学解释力： 分析 Adapter 在不同脑区（如早期视觉区 vs. 高级视觉通路）的权重分布，验证 Few-shot 学习是否主要修正了具有高个体差异的脑区（如 PPA、FFA）

\section{Experiments}
% In this section, we first describe the implementation details of MindAdapter, and then report quantitative and qualitative results on cross-subject fMRI-to-image reconstruction as well as brain functional alignment analysis. Additional metrics, ablation studies, efficiency analysis, and supplementary visualizations are provided in the Appendix.
\subsection{Implementation Details}
% The BTM is composed of two bias-free linear layers with a hidden size $h = 4096$. The input dimension \( n \) and output dimension \( k \) of BTM are determined by the specific subject transfer pairs.  
% For subjects 1, 2, 5, and 7, the dimensions are 15,724, 14,278, 13,039, and 12,682, respectively.
% The cross-stimulus neural mapper is implemented using the Feature-wise Linear Modulation model~\cite{film}, where the input dimension of \( \mathcal{M}_{\text{diff}} \) is \( a = 768 \), matching the CLIP embedding dimension, and the output is $2h = 8192$.  
% The functional embedder is a linear layer with input and output sizes of \( h = 4096 \).
% % loss
% The loss coefficients are set to $\alpha_{rec} = 1$, $\alpha_{la} =\alpha_{KL} = 0.001$, $\alpha_1 = 0.033$, and $\alpha_2 = 0.016$. The learning rates for the brain transfer matrix, cross-stimulus neural mapper, and functional embedder are all set to $1\text{e}{-5}$.
% We use a batch size of 16 and optimize using Adam. Training on a single NVIDIA A100 GPU achieves convergence in approximately 12 hours.
For subjects 1, 2, 5, and 7, the voxel dimensions are 15,724, 14,278, 13,039, and 12,682.
On top of the frozen BTM, we attach a lightweight nonlinear residual adapter implemented as a three-layer multilayer perceptron with hidden dimension 4,096 and GELU activations.
% The calibrated mapper is formulated as
% \begin{equation}
% F_\theta(\mathbf{x}) = F_0(\mathbf{x}) + A_\theta(F_0(\mathbf{x})),
% \end{equation}
% where $F_0$ denotes the pretrained BTM and $A_\theta$ denotes the trainable adapter.
A pretrained and frozen brain-to-CLIP visual decoder is employed to project predicted target-subject features into the CLIP embedding space of dimension 768 for semantic supervision.
Training is driven by a dual-stream objective consisting of an anchor-based voxel-level alignment loss and a semantic consistency loss. The weights $\alpha=2$, $\lambda=5$.
We optimize only the adapter parameters using Adam with a learning rate of $1\times10^{-4}$, while keeping the BTM and visual decoder frozen. 
The semantic stream uses a batch size of 128, while the anchor stream uses a small batch size capped at 32.
All experiments are conducted on a single NVIDIA H100 GPU, and few-shot calibration typically converges within several minutes.

% • 训练时间与参数量 (Training Cost):
%     ◦ 对比各方法的训练时间（Time per subject）和可训练参数量（Trainable Parameters）。
%     ◦ 亮点： 强调您的 Adapter 训练极快（可能只需几分钟），适合临床实时部署，而 MindEye2 可能需要更久。

% • 特征空间可视化 (t-SNE/UMAP):
%     ◦ 可视化对齐前后的特征分布。
%     ◦ 展示 32 个 Anchor 样本如何像“钉子”一样，将新被试的特征流形“拉”向标准空间，同时保持了拓扑结构不崩塌。

\subsection{Dataset and Metrics}
% We conduct all experiments on the Natural Scenes Dataset (NSD)~\cite{nsd}, a large-scale fMRI benchmark widely adopted for brain visual decoding. NSD contains neural responses from multiple subjects viewing natural images drawn from the MSCOCO-2017 dataset~\cite{coco}. Following the data-limited experimental protocol of MindEye2, we use only a single scanning session (approximately one hour of fMRI data) for training the pretrained cross-subject mapper. Few-shot calibration in MindAdapter is performed using a small number of shared stimuli between subjects, while all remaining data are treated as unpaired.

\begin{figure*}[t!]
\centering
% \vspace{-1.5em}
\includegraphics[width=0.95\textwidth]{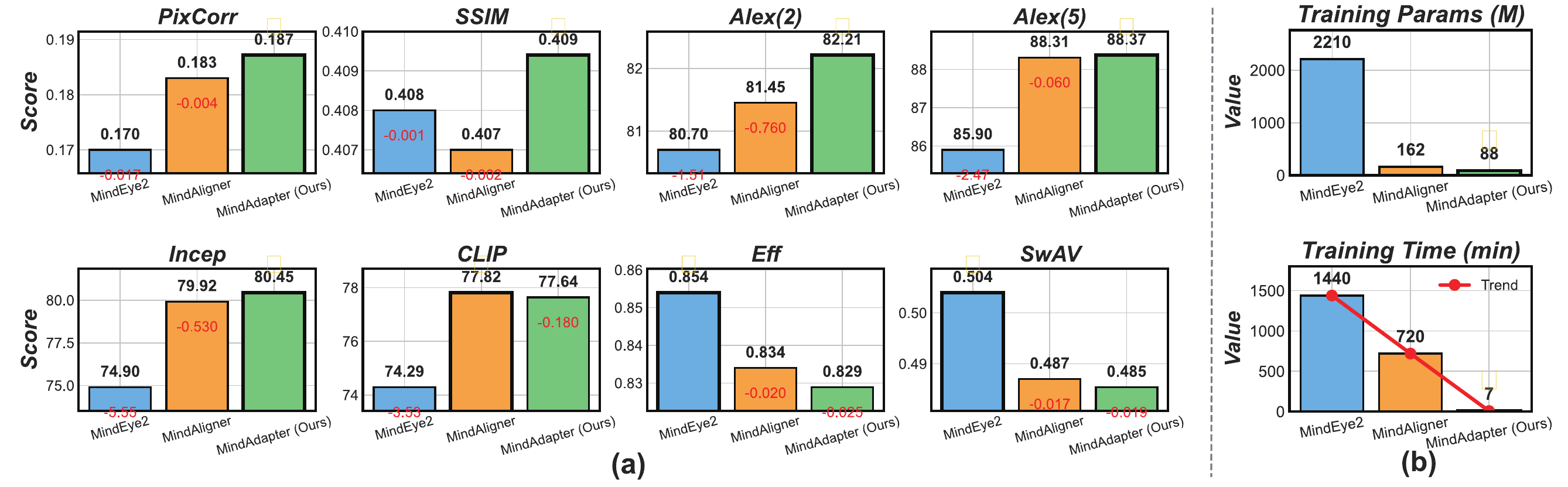}
\vspace{-1.8em}
\caption{
(a) Comparison of MindEye2, MindAligner, and MindAdapter on low-level image fidelity metrics and high-level semantic similarity metrics for cross-subject visual decoding. Red numbers indicate relative differences with respect to best.
(b) Parameter efficiency and training cost comparison.
MindAdapter requires substantially fewer trainable parameters and dramatically less training time than MindEye2 and MindAligner, enabling efficient few-shot calibration.
} 
\label{fig:method_compare_mindye2}
\vspace{-1em}
\end{figure*}

We conduct all experiments on the Natural Scenes Dataset (NSD)~\cite{nsd}, a large-scale fMRI benchmark widely adopted for brain visual decoding. NSD contains neural responses from multiple subjects viewing natural images drawn from the MSCOCO-2017 dataset~\cite{coco}. 
Following the data-limited experimental protocol of MindEye2, BTM is trained using a single scanning session (approximately one hour of fMRI data) per subject \cite{daimindaligner}. 
Few-shot calibration in MindAdapter is performed using a small number of shared stimuli between subjects, while all remaining data are treated as unpaired. 
For evaluation, we adopt two test splits. Unless otherwise specified, we report quantitative and qualitative results on a test set containing 740 images. In addition, we use a larger hold-out test set with 1,000 images for the analysis in \autoref{fig:method_compare_mindye2}.

\begin{figure}[t!]
\centering
\includegraphics[width=\columnwidth]{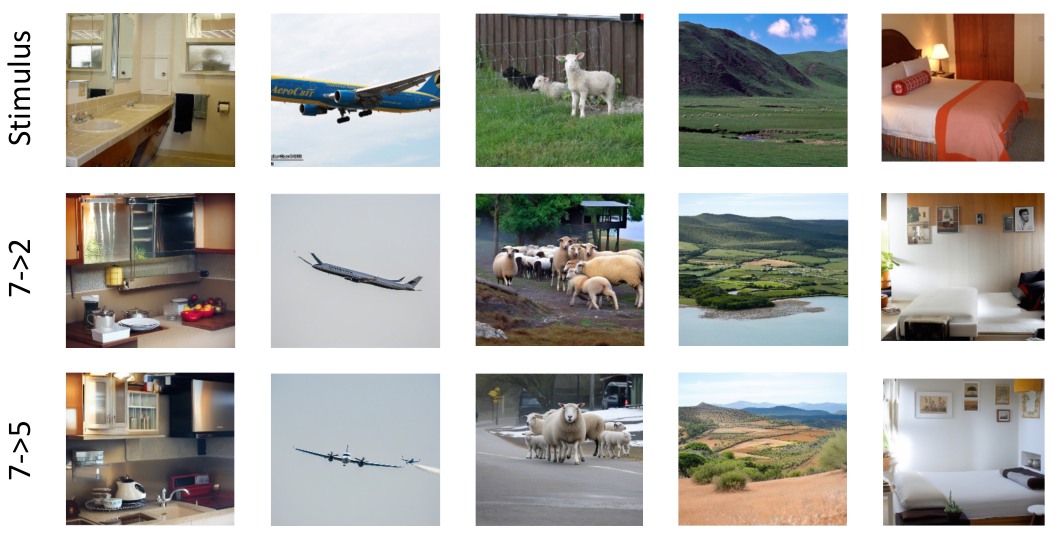}
\vspace{-2em}
\caption{
% Few-shot scaling comparison between the few-shot variant of MindAligner and MindAdapter across multiple metrics.
% Results show that naive incorporation of few-shot anchor supervision into MindAligner is ineffective, whereas MindAdapter achieves reliable few-shot adaptation.
Effect of aligning a fixed target subject with different source subjects.
MindAdapter yields consistent reconstructions across alignment directions, indicating robust and source-invariant cross-subject calibration.
} 
\label{fig:vis2}
\vspace{-1.5em}
\end{figure}

To evaluate fMRI-to-image reconstruction performance, we assess both low-level visual fidelity and high-level semantic consistency of the reconstructed images. Low-level metrics capture basic visual properties such as pixel-wise similarity and edge structures, whereas high-level metrics reflect semantic correspondence between reconstructions and ground-truth images. Following prior works~\cite{mindeyev1, mindeyev2}, we adopt PixCorr, SSIM, AlexNet(2), and AlexNet(5)~\cite{alex} to measure low-level fidelity, and use Inception~\cite{inception}, CLIP~\cite{clip}, EffNet-B~\cite{effi}, and SwAV~\cite{swav} to evaluate high-level semantic similarity. All metrics are computed by comparing reconstructed images against their ground-truth counterparts \cite{daimindaligner}.

% To evaluate functional alignment, we use two metrics: fMRI Spatial Correlation (fSC)~\cite{corr} and Transfer Quantity (TQ). fSC measures the Pearson correlation between corresponding brain regions of two subjects ($i \neq j$), assessing global alignment consistency.
% TQ captures voxel-level differences by analyzing the weights of the BTM $\boldsymbol{\mathcal{M}}$, which maps voxels between subjects. For a source voxel indexed by $i$, TQ is defined as $\text{TQ}_i =\sum_{0 \leq j < p} ||\boldsymbol{\mathcal{M}}_{i, j}||$, where $p$ is the number of voxels in the target brain. High TQ values indicate regions with greater activation differences and more intricate functional alignment requirements.

\subsection{Main Results}

% We evaluate the visual decoding performance of MindAligner and the proposed MindAdapter using both qualitative and quantitative analyses. We compare against representative state-of-the-art brain-to-image decoding frameworks, including MindEye2~\cite{mindeyev2} and MindAligner~\cite{daimindaligner}.

We evaluate the proposed MindAdapter and compare it with state-of-the-art brain-to-image decoding methods, including MindEye2~\cite{mindeyev2} and MindAligner~\cite{daimindaligner}.

% \begin{figure}[t!]
% \centering
% \includegraphics[width=\columnwidth]{Figure/Figure7.png}
% \vspace{-1.5em}
% \caption{
% Parameter efficiency and training cost comparison.
% MindAdapter requires substantially fewer trainable parameters and dramatically less training time than MindEye2 and MindAligner, enabling efficient few-shot calibration.
% } 
% \label{fig:time}
% \vspace{-2em}
% \end{figure}

\noindent \textbf{Quantitative Comparison.}
\autoref{tab:64shot_comparison} reports quantitative results under the 64-shot setting. 
MindAdapter consistently outperforms MindAligner across both low-level and high-level metrics on all evaluated subjects. 
For instance, on Subject~1, MindAdapter improves PixCorr from 0.2136 to 0.2239, AlexNet(2) accuracy from 87.64\% to 88.43\%, and Inception accuracy from 83.19\% to 84.30\%, indicating enhanced structural fidelity and semantic consistency. 
Similar improvements are observed across other subjects.
\autoref{fig:method_compare_mindye2} (a) further compares MindAdapter with MindEye2 and MindAligner. 
While MindAligner already surpasses MindEye2 via explicit cross-subject functional alignment, MindAdapter achieves additional gains across nearly all metrics, demonstrating that the nonlinear residual adapter effectively corrects subject-specific deviations beyond linear transfer.
\autoref{fig:fewshotmetric_1} analyzes the effect of calibration shots. 
Performance improves monotonically as the number of anchors increases, with strong results already achieved using 32--64 shots, highlighting the data efficiency of the proposed coarse-to-fine calibration paradigm.

\noindent \textbf{Qualitative Comparison.}
% {\autoref{vis1}} presents qualitative comparisons between MindAligner and our MindAdapter under the single-session setting.
% While MindAligner is able to recover coarse semantic content, its reconstructions often exhibit imprecise object localization and missing fine-grained structures.
% In contrast, MindAdapter produces reconstructions with clearer object boundaries, more accurate spatial layouts, and improved semantic fidelity, demonstrating that few-shot residual calibration effectively corrects subject-specific distortions without disrupting the pretrained alignment space.
% \textcolor{red}{\autoref{fig:vis3}} further visualizes the effect of the number of calibration shots on reconstruction quality.
% As the number of anchor samples increases from 32 to 256, the reconstructed images progressively become sharper and more semantically consistent with the stimuli, with clearer object shapes and reduced artifacts.
% This monotonic visual improvement highlights the strong data efficiency of MindAdapter and confirms that even a small number of shared stimuli can substantially enhance cross-subject visual decoding.
\autoref{vis1} presents qualitative comparisons between MindAligner and MindAdapter under the single-session setting. 
While MindAligner recovers coarse semantics, it often suffers from imprecise localization and missing fine structures. 
In contrast, MindAdapter yields more accurate spatial layouts, and higher semantic fidelity, indicating that few-shot residual calibration effectively corrects subject-specific distortions without disturbing the pretrained alignment.
\autoref{fig:vis3} further shows the impact of calibration shots. 
As the number of anchors increases from 32 to 256, reconstructions become more semantically consistent, demonstrating the strong data efficiency of MindAdapter.

% Fig.~\ref{fig:1hvis} visualizes representative fMRI-to-image reconstruction results obtained using a single fMRI session (approximately one hour of data). 
% Compared with MindEye2 and MindAligner, MindAdapter produces reconstructions with clearer structures, sharper boundaries, and more accurate semantic content.
% These qualitative improvements are consistent with the quantitative results in \textcolor{red}{\autoref{tab:64shot_comparison}}, where MindAdapter achieves consistent gains over MindAligner across both low-level fidelity and high-level semantic metrics for multiple subjects. 
% In addition, \textcolor{red}{\autoref{fig:fewshotmetric_1}} shows that MindAdapter scales favorably as the number of few-shot anchor samples increases, demonstrating effective utilization of limited paired supervision. 
% As further evidenced in \textcolor{red}{\autoref{fig:method_compare}}, MindAdapter outperforms MindEye2 and MindAligner across a wide range of evaluation metrics.
% Overall, these results indicate that the proposed few-shot residual calibration effectively injects subject-specific corrections into a frozen cross-subject alignment, enabling accurate transfer of the decoding model under limited data.

\subsection{Ablation and Analysis Study}
% To evaluate the effectiveness of each model design in MindAligner, we perform an ablation study using Subject 2 as the novel subject and Subject 1 as the known subject. The results exclude the refinement step of MindEye2 for generated images. As shown in Tab.~\ref{tab:ablation_study}, training MindAligner with only the visual decoding loss $\boldsymbol{\mathcal{L}}_{dec}$ yields suboptimal cross-subject reconstruction performance, underscoring the difficulty of directly generalizing pre-trained models to new subjects without effective alignment.
% Adding signal reconstruction loss $\boldsymbol{\mathcal{L}}_{rec}$ significantly enhances performance as it leads to more accurate brain activity reconstructions.
% The incorporation of $\boldsymbol{\mathcal{L}}_{KL}$ further strengthens alignment by enforcing consistency between the distributions of the generated and real signals. 
% Lastly, $\boldsymbol{\mathcal{L}}_{latent}$ exploits the correlation of visual stimuli and fMRI embeddings to guide the brain alignment in the latent space, thereby improving model's ability to capture visual semantics in brain activity and enhancing low-level reconstruction performance. These losses together work in synergy to refine alignment and improve cross-subject decoding fidelity.
\noindent \textbf{Ablation Study.}
We conduct an ablation study to examine the contribution of each component in MindAdapter under a fixed cross-subject transfer setting.
As shown in \autoref{tab:ablation_study}, removing anchor supervision ($\mathcal{L}_{\text{Anchor}}$) leads to a substantial performance drop across all metrics, indicating that few-shot paired anchors are indispensable for subject-specific calibration.
Excluding the anchor MSE term while retaining contrastive supervision (w/o $\mathcal{L}_{\text{Anchor}}^{\text{MSE}}$) also degrades performance, highlighting the stabilizing role of voxel-level regression in residual correction.
Removing the semantic consistency loss $\mathcal{L}_{\text{sec}}$ consistently harms performance, demonstrating the necessity of unpaired semantic guidance for preserving visual semantics during adaptation.
Moreover, discarding the coarse stage (w/o CoarseStage), i.e., directly adapting without the frozen BTM, causes severe performance collapse, confirming that MindAdapter relies on coarse cross-subject correspondence and functions as a residual calibrator rather than a replacement.
Finally, substituting the nonlinear adapter with a linear variant (w/ LinearAdapter) results in inferior performance, underscoring the importance of nonlinear residual modeling.
Overall, the full model achieves the best results, validating the effectiveness of the proposed anchor supervision, semantic regularization, coarse-to-fine design, and nonlinear residual adaptation.

% \subsubsection{Shot number}

% \begin{table*}[!t]
% \centering
% \caption{Ablation study results. The combination of $\boldsymbol{\mathcal{L}}_{dec}$+$\boldsymbol{\mathcal{L}}_{rec}$+$\boldsymbol{\mathcal{L}}_{KL}$+$\boldsymbol{\mathcal{L}}_{latent}$ is our final model setting.}
% \resizebox{0.78\textwidth}{!}{
% \begin{tabular}{@{}lcccccccc@{}}
% \toprule
% \textbf{Method} & \textbf{PixCorr$\uparrow$} & \textbf{SSIM$\uparrow$} & \textbf{Alex(2)$\uparrow$} & \textbf{Alex(5)$\uparrow$} & \textbf{Incep$\uparrow$} & \textbf{CLIP$\uparrow$} & \textbf{Eff$\downarrow$} & \textbf{SwAV$\downarrow$} \\
% \midrule
% +$\boldsymbol{\mathcal{L}}_{dec}$ & 0.072 & 0.318 & 63.50\% & 71.44\% & 66.07\% & 62.59\% & 0.905 & 0.550 \\
% +$\boldsymbol{\mathcal{L}}_{dec}$+$\boldsymbol{\mathcal{L}}_{rec}$ & 0.186 & 0.340 & 86.83\% & 93.51\% & 84.55\% & 82.42\% & 0.811 & 0.465 \\
% +$\boldsymbol{\mathcal{L}}_{dec}$+$\boldsymbol{\mathcal{L}}_{rec}$+$\boldsymbol{\mathcal{L}}_{KL}$ & 0.191 & 0.407 & 87.98\% & 92.99\% & \textbf{86.61\%} & 82.16\% & \textbf{0.780} & \textbf{0.453} \\
% +$\boldsymbol{\mathcal{L}}_{dec}$+$\boldsymbol{\mathcal{L}}_{rec}$+$\boldsymbol{\mathcal{L}}_{KL}$+$\boldsymbol{\mathcal{L}}_{latent}$ & \textbf{0.195} & \textbf{0.408} & \textbf{88.25\%} & \textbf{93.51\%} & 86.24\% & \textbf{82.72\%} & 0.782 & 0.454 \\
% \bottomrule
% \end{tabular}%
% }
% \label{tab:ablation_study}
% \vspace{-3mm}
% \end{table*}

\begin{table*}[!t]
\centering
\caption{
Ablation of MindAdapter components.
Removing anchor supervision, semantic loss, coarse stage, or nonlinear adapter degrades performance, while the full model achieves the best overall results, validating the necessity of each component.
}
\vspace{-1em}
\resizebox{0.7\textwidth}{!}{
\begin{tabular}{@{}lcccccccccc@{}}
\toprule
\textbf{Method} 
& \textbf{PixCorr$\uparrow$} 
& \textbf{SSIM$\uparrow$} 
& \textbf{Alex(2)$\uparrow$} 
& \textbf{Alex(5)$\uparrow$} 
& \textbf{Incep$\uparrow$} 
& \textbf{CLIP$\uparrow$} 
& \textbf{Eff$\downarrow$} 
& \textbf{SwAV$\downarrow$} 
& \textbf{Fwd$\uparrow$} 
& \textbf{Bwd$\uparrow$} \\
\midrule

$w/o \ \ \boldsymbol{\mathcal{L}}_{Anchor}$
& .1309 & .3953 & .8067 & .8497 & .7601 & .7098 & .8732 & .5398 & .7351 & .6348 \\

$w/o \ \ \boldsymbol{\mathcal{L}}_{Anchor}^{MSE}$ 
& .1956 & .4156 & .8399 & .8863 & .7762 & .7559 & .8501 & .5012 & .7841 & .6962 \\

$w/o \ \ \ \boldsymbol{\mathcal{L}}_{sec}$
& .2177 & .4164 & .8822 & .9289 & .8426 & .8256 & .7898 & .4511 & .9110 & .8664 \\

$w/o \ \ \  CoarseStage$
& .1172 & .3760 & .7252 & .7949 & .6962 & .6987 & .8980 & .5341 & .3171 & .2487 \\

$w/ \ \ \ Linear Adapter $  
& .2049 & .4184 & .8801 & .9326 & .8434 & .8239 & .7887 & .4504 & .3118 & .2484 \\

\midrule
\rowcolor{gray!12}
\textbf{Full Model (Ours)} 
& \textbf{.2183} & \textbf{.4164} & \textbf{.8826} & \textbf{.9330} 
& \textbf{.8436} & \textbf{.8229} & \textbf{.7889} & \textbf{.4504} 
& \textbf{.9171} & \textbf{.8707} \\
\bottomrule
\end{tabular}}
\label{tab:ablation_study}
\vspace{-3mm}
\end{table*}

\begin{figure}[t!]
\centering
\includegraphics[width=\columnwidth]{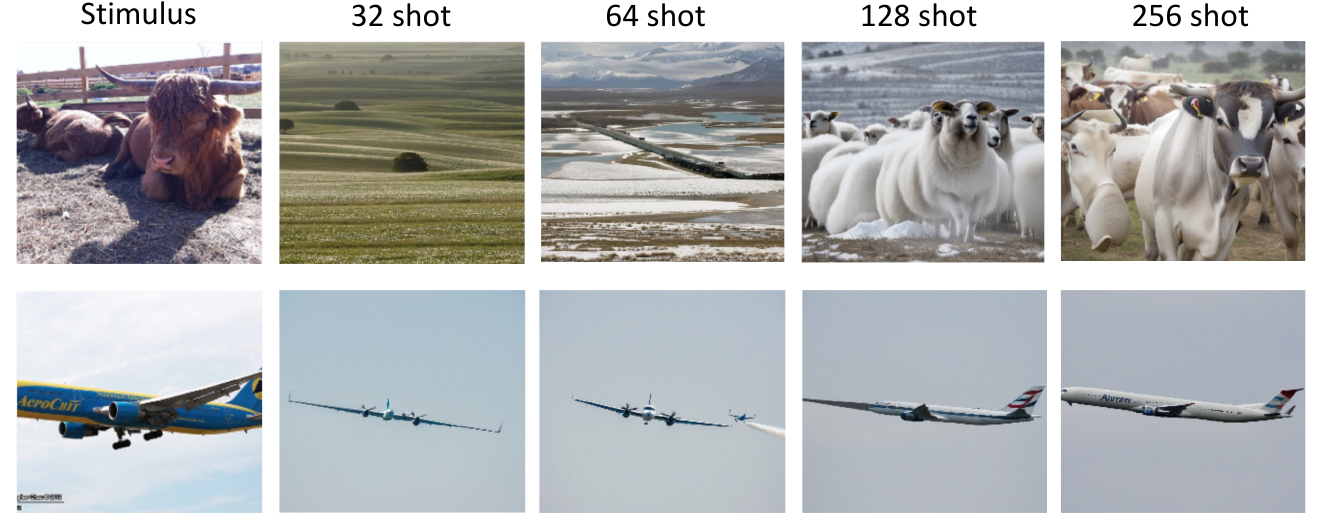}
\vspace{-2.4em}
\caption{
% Effect of shot number on reconstruction quality.
% From left to right, reconstructions progressively improve as the number of calibration shots increases (32→256), exhibiting clearer object structures, more accurate semantics, and reduced artifacts, demonstrating the strong data-efficiency and monotonic performance gain of MindAdapter.
Effect of shot number on reconstruction quality.
As the number of calibration shots increases (32$\rightarrow$256), reconstructions become more semantically faithful. 
% demonstrating MindAdapter’s strong data efficiency and monotonic gains.
} 
\label{fig:vis3}
\vspace{-2.1em}
\end{figure}

\noindent \textbf{Effect of Shot Number.}
% We investigate how the number of few-shot anchor samples influences the calibration and reconstruction performance of MindAdapter.
% As illustrated in \textcolor{red}{\autoref{fig:vis3}}, increasing the number of calibration shots from 32 to 256 leads to progressively improved visual quality, with clearer object structures, more accurate semantics, and fewer visual artifacts.
% This trend indicates that MindAdapter can effectively leverage additional anchor supervision to refine subject-specific residual corrections.
% The quantitative results in \textcolor{red}{\autoref{fig:fewshotmetric_1}} further demonstrate a consistent and monotonic performance gain across low-level image fidelity metrics (PixCorr, SSIM, AlexNet(2/5)), high-level semantic similarity metrics (Inception, CLIP, EffNet-B, SwAV), and retrieval accuracy (forward and backward).
% Notably, substantial improvements are already observed with a small number of shots (e.g., 32 or 64), highlighting the strong data efficiency of MindAdapter, while additional shots continue to yield stable incremental gains.
% These observations validate the effectiveness of the proposed few-shot calibration mechanism and its ability to smoothly scale with increasing supervision.
% Additional metric curves and analyses for different transfer directions are provided in \textcolor{red}{\autoref{fig:fewshotmetric_2}},\textcolor{red}{\autoref{fig:fewshotmetric_3}} and \textcolor{red}{\autoref{fig:fewshotmetric_4}}.
We analyze the effect of the number of few-shot anchor samples on the calibration and reconstruction performance of MindAdapter.
As shown in \autoref{fig:vis3}, increasing the number of calibration shots from 32 to 256 progressively improves visual quality, yielding clearer object structures, and more accurate semantics.
Consistent with these observations, \autoref{fig:fewshotmetric_1} reports monotonic performance gains across low-level fidelity metrics (PixCorr, SSIM, AlexNet(2/5)), high-level semantic metrics (Inception, CLIP, EffNet-B, SwAV), and retrieval accuracy.
Notably, strong improvements are already achieved with as few as 32--64 shots, demonstrating the high data efficiency of MindAdapter, while additional anchors provide stable incremental benefits.
These results confirm the effectiveness and scalability of the proposed few-shot calibration mechanism.
More analyses for different transfer directions are provided in \autoref{fig:fewshotmetric_2}, \autoref{fig:fewshotmetric_3}, and \autoref{fig:fewshotmetric_4}.

\noindent \textbf{Balance Between Anchor and Semantic Streams.}
We analyze the impact of balancing anchor-based and semantic consistency losses in {\autoref{fig:dualstreamparas}}.
Performance is stable across a broad range of weights, with moderate balancing achieving the best PixCorr.
This demonstrates that the two streams are complementary: the anchor stream enables subject-specific calibration, while the semantic stream preserves global semantic structure, and both are necessary for optimal performance.

\noindent \textbf{Comparison with Few-Shot Variant of MindAligner.}
\textcolor{red}{\autoref{fig:comparefewshotMindAligner}} compares MindAdapter with a naive few-shot extension of MindAligner under increasing numbers of shared anchor samples.
Directly injecting few-shot anchor supervision into MindAligner yields limited and unstable improvements across both low-level fidelity and high-level semantic metrics, indicating that linear alignment alone is insufficient for effective subject-specific calibration.
In contrast, MindAdapter consistently achieves higher performance and exhibits clear scaling behavior as the number of shots increases, validating the necessity of nonlinear residual adaptation and dual-stream supervision for reliable few-shot cross-subject transfer.

\noindent \textbf{Impact of Aligning to Different Subjects.}
% We visualize the reconstruction results of fixing a target subject and aligning it with different source subjects under few-shot settings in \textcolor{red}{\autoref{fig:vis2}}.
% MindAdapter produces visually consistent reconstructions across different source--target pairs, while the few-shot variant of MindAligner exhibits noticeable degradation and instability.
% This indicates that MindAdapter achieves robust cross-subject calibration regardless of which source subject is used.
% The robustness arises from MindAdapter’s residual nonlinear calibration on top of a frozen cross-subject mapper, which enables subject-specific correction while preserving shared functional structure.
% In contrast, directly injecting few-shot anchor supervision into MindAligner fails to establish reliable calibration.
We examine the robustness of MindAdapter by fixing a target subject and aligning it with different source subjects under the few-shot setting.
As shown in \autoref{fig:vis2}, MindAdapter produces visually consistent and semantically stable reconstructions across different alignment directions (e.g., $7\!\rightarrow\!2$ and $7\!\rightarrow\!5$).
This demonstrates that the proposed residual calibration is insensitive to the choice of source subject and enables reliable cross-subject adaptation.

\noindent \textbf{Computational Efficiency.}
We compare the computational efficiency of MindAdapter with MindEye2 and MindAligner in terms of trainable parameters and training time, as shown in \autoref{fig:method_compare_mindye2} (b).
MindAdapter introduces only a lightweight nonlinear adapter on top of a frozen cross-subject mapper, resulting in substantially fewer trainable parameters than both MindEye2 and MindAligner.
Specifically, MindAdapter uses 88M trainable parameters, compared to 162M for MindAligner and 2210M for MindEye2.
In addition, MindAdapter requires dramatically less training time, completing few-shot calibration within minutes, whereas MindAligner requires several hours and MindEye2 requires more than one day of training.
These results demonstrate that MindAdapter achieves high decoding performance with minimal computational overhead, making it well-suited for practical few-shot cross-subject calibration.

\noindent \textbf{t-SNE and ROI Visualization.}
% We further analyze the representation structure and functional organization induced by MindAdapter through embedding visualization and surface-based ROI analysis. As shown in \textcolor{red}{\autoref{fig:tsne}}, compared to MindAligner, MindAdapter produces more compact and better-separated semantic clusters in the embedding space \cite{maaten2008visualizing}, indicating reduced representation dispersion and improved cross-subject alignment after few-shot calibration. This suggests that the nonlinear residual adapter effectively refines coarse cross-subject correspondence into a more discriminative and semantically organized latent space.  
We further examine the representational structure induced by MindAdapter through embedding visualization and surface-based ROI analysis.
As shown in \textcolor{red}{\autoref{fig:tsne}}, compared to MindAligner, MindAdapter yields more compact and better-separated semantic clusters in the embedding space~\cite{maaten2008visualizing}, indicating reduced representation dispersion and improved cross-subject alignment after few-shot calibration.
These results suggest that the residual adapter effectively refines coarse cross-subject correspondence into a more discriminative and semantically organized latent space.
% In addition, \textcolor{red}{\autoref{fig:roi_vis}} visualizes surface-based ROI-level importance maps derived from calibrated subject-specific models and projected onto a common cortical surface. Despite subject-specific calibration, the learned importance patterns exhibit strong spatial consistency across subjects, revealing a shared model-derived functional topology. These results demonstrate that MindAdapter preserves global functional organization while injecting individualized residual corrections, leading to improved alignment without disrupting large-scale functional structure.
In addition, \textcolor{red}{\autoref{fig:roi_vis}} visualizes surface-based ROI importance maps derived from subject-specifically calibrated models and projected onto a common cortical surface.
Despite individualized calibration, the learned importance patterns remain highly consistent across subjects, revealing a shared model-induced functional topology.
These results demonstrate that MindAdapter injects subject-specific residual corrections while preserving large-scale functional organization, achieving improved alignment without disrupting global brain structure.

\section{Limitations and Ethical Considerations}

% \noindent \textbf{Limitations.}
Despite the clear improvements brought by MindAdapter, several limitations remain.
As illustrated in \textcolor{red}{\autoref{fig:vis4}}, while MindAdapter generally preserves the dominant color tone and coarse scene semantics, it can still struggle with fine-grained object geometry, complex spatial layouts, and multi-object composition, resulting in structural distortions or mismatched details in challenging scenes.
In addition, although MindAdapter is data-efficient, reliable calibration typically requires at least 16 shared anchor stimuli between subjects to achieve stable performance, and extremely low-shot regimes may lead to suboptimal adaptation.
Addressing these limitations will require stronger structural priors and more powerful generative decoders in future work.

This study uses only publicly available, de-identified datasets released for research use. No new data are collected, and no personally identifiable information is accessed. Informed consent was obtained by the original data collectors, and the study involves no direct interaction with human participants.

\section{Conclusion}
We propose \textit{MindAdapter}, a parameter-efficient few-shot calibration framework for cross-subject brain-to-visual decoding.
MindAdapter introduces a \emph{decoupled linear--residual cascade} that freezes a linear cross-subject alignment backbone and augments it with a lightweight nonlinear residual adapter, enabling coarse-to-fine subject-specific calibration without retraining large models.
To overcome extreme data scarcity, we further design a \emph{topology-anchored dual-stream objective} that integrates voxel-level paired anchors with unpaired semantic regularization in the visual embedding space, ensuring stable optimization and preserved functional topology.
Extensive experiments on the NSD show that MindAdapter consistently improves cross-subject reconstruction and retrieval few-shot shared stimuli while remaining computationally efficient.
These results establish MindAdapter as a practical and scalable solution for personalized brain decoding under realistic data constraints.

\section*{GenAI Disclosure}

Generative AI tools were used in a limited and auxiliary manner during the preparation of this manuscript. 
Specifically, large language models were employed to assist with language polishing, clarity improvement, and structural refinement of the presentation. 
All scientific ideas, methodological designs, experimental implementations, results, and conclusions were conceived, conducted, and verified by the authors. 
No generative AI tools were used for data collection, data analysis, experiment execution, result interpretation, or decision-making in the research process. 
The authors take full responsibility for the content of this paper.

% Generative AI tools were used solely to assist with language editing and presentation clarity during the preparation of this manuscript. 
% They did not contribute to the research design, methodology, experimental implementation, data analysis, or interpretation of results. 
% All technical contributions and conclusions are the sole work of the authors.

\bibliographystyle{ACM-Reference-Format}
\bibliography{sample-base}

\newpage
\appendix

\section{Metrics}
Following prior work~\cite{mindeyev1,mindeyev2,daimindaligner}, we evaluate the image reconstruction results based on eight metrics, which are categorized into low-level and high-level groups. Low-level metrics, including Pixelwise Correlation (PixCorr), Structural Similarity Index (SSIM)~\cite{ssmi}, AlexNet(2) (Alex(2)), and AlexNet(5) (Alex(5))~\cite{alex}, focus on textural and structural details. High-level metrics—Inception (Incep)~\cite{inception}, CLIP~\cite{clip}, EfficientNet-B (Eff)~\cite{effi}, and SwAV-ResNet50 (SwAV)~\cite{swav} —assess semantic fidelity. Alex(2), Alex(5), Incep, and CLIP metrics are derived by calculating Pearson correlation between the embeddings of the ground truth and reconstructed images, following the two-way identification framework of Ozcelik and VanRullen~\cite{braindiffuser}. Eff and SwAV scores are based on the average distance between feature embeddings.

In addition to reconstruction-based metrics, we evaluate cross-subject decoding performance using retrieval-based metrics to quantify fine-grained semantic alignment in the visual embedding space, following the evaluation protocol of MindEye2~\cite{mindeyev2}. 
Specifically, each predicted brain-derived representation is compared against CLIP image embeddings from 300 randomly sampled images in the test set using cosine similarity. 
\emph{Forward retrieval accuracy} (\textbf{fwd\_corr}) measures whether a predicted brain representation correctly retrieves its corresponding ground-truth image as the top-1 match (with random chance at $1/300$). 
Conversely, \emph{backward retrieval accuracy} (\textbf{bwd\_corr}) evaluates the inverse direction, assessing whether a given image embedding retrieves its corresponding predicted brain representation among all candidates.
To reduce variance induced by random sampling, the retrieval process is repeated 30 times per test sample and the results are averaged.
Together, these bidirectional retrieval metrics provide a robust assessment of alignment quality between brain-derived representations and visual semantics, capturing both discriminability and cross-modal consistency of the learned mapping.

\section{Additional Analysis on Few-Shot Adaptation}
\label{app:fewshot_mindaligner}

\begin{figure*}[t!]
\centering
\includegraphics[width=\textwidth]{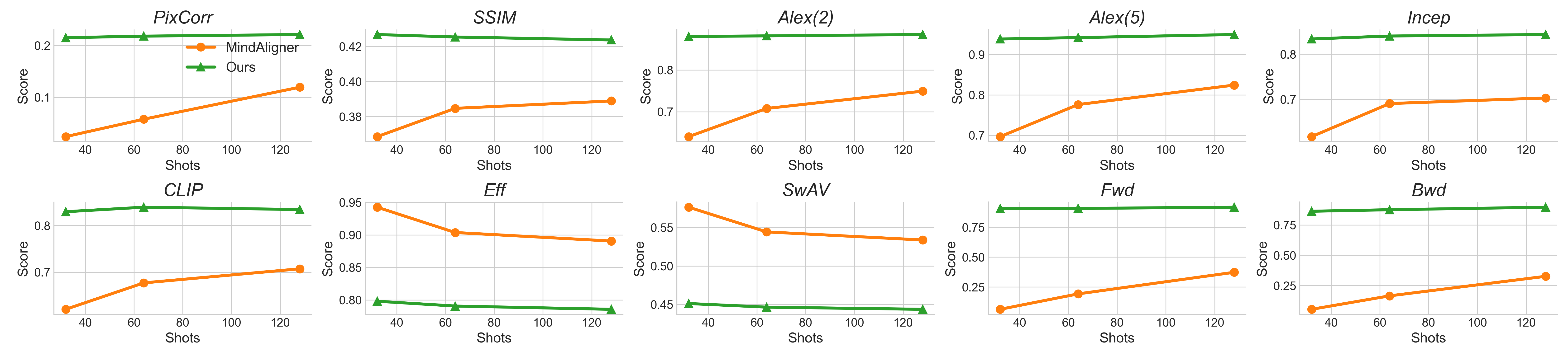}
% \vspace{-1.5em}
\caption{
Few-shot scaling comparison between the few-shot variant of MindAligner and MindAdapter across multiple metrics (2->1).
Results show that naive incorporation of few-shot anchor supervision into MindAligner is ineffective, whereas MindAdapter achieves reliable few-shot adaptation.
} 
\label{fig:comparefewshotMindAligner}
\vspace{-1.5em}
\end{figure*}

\begin{figure*}[t!]
\centering
\includegraphics[width=\textwidth]{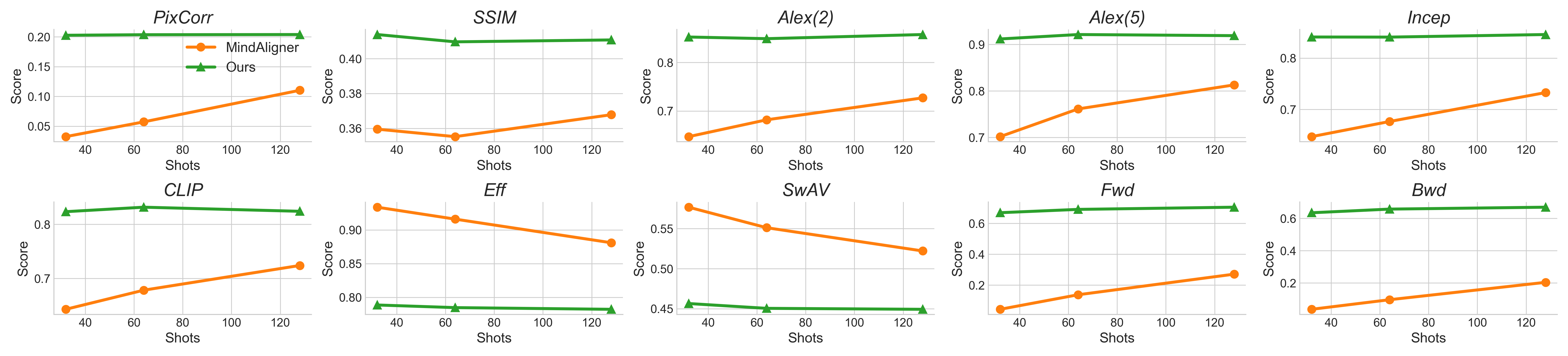}
% \vspace{-1.5em}
\caption{
Few-shot scaling comparison between the few-shot variant of MindAligner and MindAdapter across multiple metrics (5->1).
Results show that naive incorporation of few-shot anchor supervision into MindAligner is ineffective, whereas MindAdapter achieves reliable few-shot adaptation.
} 
\label{fig:comparefewshotMindAligner5to1}
\vspace{-1.5em}
\end{figure*}

\begin{figure*}[t!]
\centering
\includegraphics[width=\textwidth]{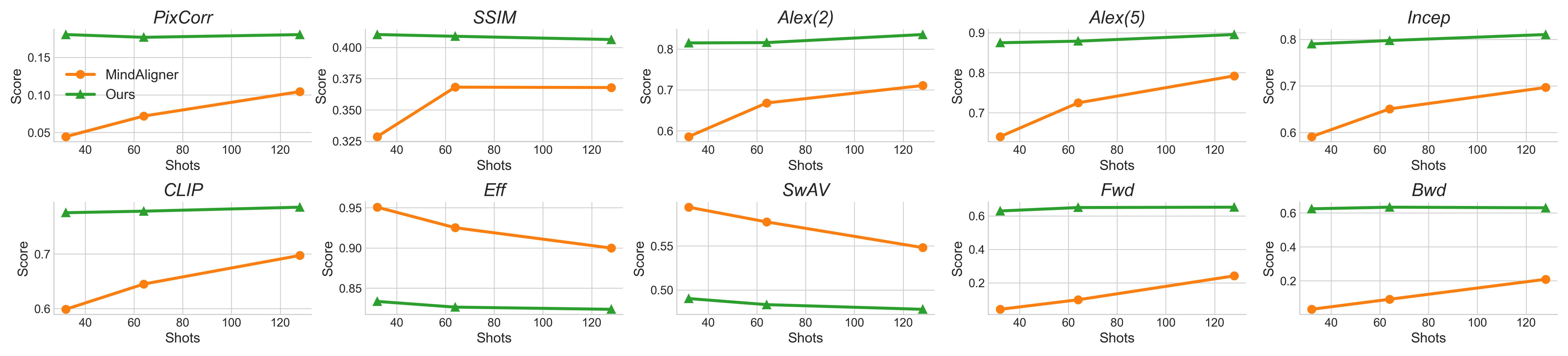}
% \vspace{-1.5em}
\caption{
Few-shot scaling comparison between the few-shot variant of MindAligner and MindAdapter across multiple metrics (7->1).
Results show that naive incorporation of few-shot anchor supervision into MindAligner is ineffective, whereas MindAdapter achieves reliable few-shot adaptation.
} 
\label{fig:comparefewshotMindAligner7to1}
\vspace{-1.5em}
\end{figure*}

\paragraph{Comparison with Few-Shot MindAligner.}
We further provide a detailed comparison between MindAdapter and a naive few-shot variant of MindAligner, where the original MindAligner is augmented with few-shot anchor supervision without modifying its linear alignment structure.
\autoref{fig:comparefewshotMindAligner}, \autoref{fig:comparefewshotMindAligner5to1}, and \autoref{fig:comparefewshotMindAligner7to1} report few-shot scaling results for three representative cross-subject transfer directions (2$\!\rightarrow\!$1, 5$\!\rightarrow\!$1, and 7$\!\rightarrow\!$1), evaluated across a broad set of low-level, high-level, and retrieval-based metrics.

Across all transfer directions and metrics, MindAdapter exhibits a clear and consistent monotonic improvement as the number of calibration shots increases.
In contrast, the few-shot variant of MindAligner shows limited gains and often unstable or saturated behavior, particularly on high-level semantic metrics (CLIP, Inception, EffNet, SwAV) and retrieval accuracy (fwd\_corr, bwd\_corr).
These results indicate that directly injecting few-shot supervision into a linear cross-subject alignment model is insufficient for reliable adaptation.

A key observation is that MindAligner relies on a single linear BTM to model cross-subject correspondence.
While effective at capturing coarse functional alignment, this linear formulation lacks the capacity to absorb subject-specific residual deviations exposed in few-shot settings.
As a result, additional anchor supervision cannot be effectively translated into semantic improvements and may even introduce local distortions in the learned representation space.
In contrast, MindAdapter explicitly adopts a coarse-to-fine calibration strategy.
The frozen BTM preserves global cross-subject correspondence, while the lightweight nonlinear residual adapter selectively corrects structured subject-specific misalignment.
This design allows few-shot anchor signals to be absorbed as targeted residual corrections, leading to smooth and stable performance scaling with increasing shots.
Notably, MindAdapter already achieves strong performance with as few as 32--64 shared stimuli and continues to improve consistently up to 128 shots across all evaluated metrics.
This behavior is observed uniformly across low-level image fidelity metrics (PixCorr, SSIM, AlexNet), high-level semantic similarity metrics (Inception, CLIP, EffNet, SwAV), and bidirectional retrieval accuracy.
These results confirm that reliable few-shot cross-subject adaptation requires explicit residual modeling rather than naive fine-tuning of linear alignment.
Overall, the comparisons in Figures~\autoref{fig:comparefewshotMindAligner}--\autoref{fig:comparefewshotMindAligner7to1} demonstrate that MindAdapter provides a principled and robust solution for few-shot cross-subject calibration, whereas few-shot extensions of MindAligner fail to fully exploit limited paired supervision.

\begin{figure}[t!]
\centering
\includegraphics[width=\columnwidth]{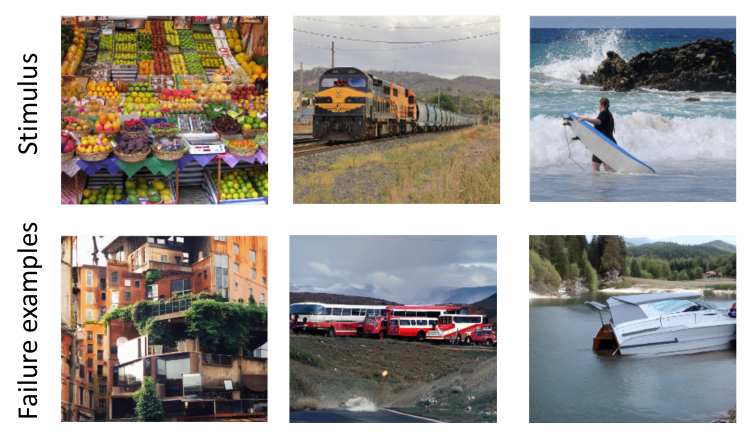}
% \vspace{-1.5em}
\caption{
Although MindAdapter preserves the dominant color tone and coarse scene semantics, it may still struggle with fine-grained object geometry, spatial layout, and multi-object composition, leading to structural distortions or mismatched details in complex scenes.
} 
\label{fig:vis4}
\vspace{-1.5em}
\end{figure}

\section{Additional Analysis on Few-Shot Scaling Behavior}
\label{app:fewshot_scaling}

\begin{figure*}[t!]
\centering
\includegraphics[width=\textwidth]{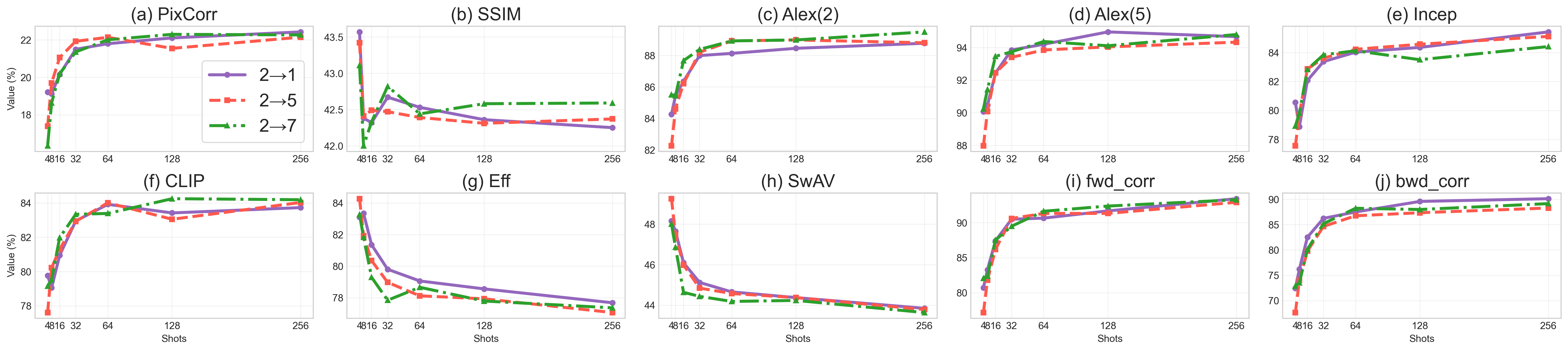}
% \vspace{-1.5em}
\caption{
Performance scaling of MindAdapter with increasing numbers of few-shot anchor samples for different cross-subject transfer directions (2→1, 2→5, 2→7), evaluated by low-level image fidelity, high-level semantic similarity, and retrieval accuracy.
} 
\label{fig:fewshotmetric_2}
\end{figure*}

\begin{figure*}[t!]
\centering
\includegraphics[width=\textwidth]{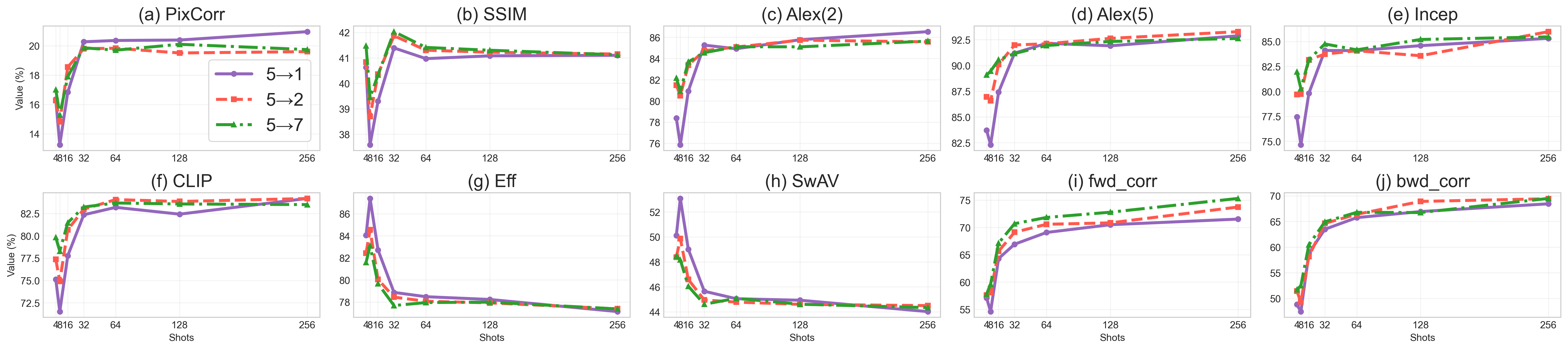}
% \vspace{-1.5em}
\caption{
Performance scaling of MindAdapter with increasing numbers of few-shot anchor samples for different cross-subject transfer directions (5→1, 5→2, 5→7), evaluated by low-level image fidelity, high-level semantic similarity, and retrieval accuracy.
} 
\label{fig:fewshotmetric_3}
\end{figure*}

\begin{figure*}[t!]
\centering
\includegraphics[width=\textwidth]{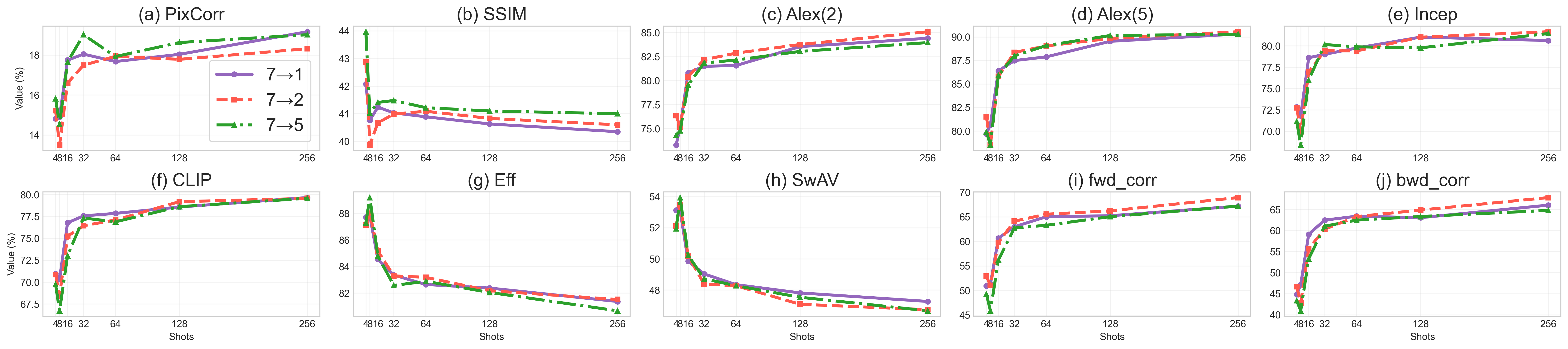}
% \vspace{-1.5em}
\caption{
Performance scaling of MindAdapter with increasing numbers of few-shot anchor samples for different cross-subject transfer directions (7→1, 7→2, 7→5), evaluated by low-level image fidelity, high-level semantic similarity, and retrieval accuracy.
} 
\label{fig:fewshotmetric_4}
\end{figure*}

We provide a detailed analysis of how the performance of \textsc{MindAdapter} scales with the number of few-shot anchor samples across different cross-subject transfer directions.
\autoref{fig:fewshotmetric_1}, \autoref{fig:fewshotmetric_2}, \autoref{fig:fewshotmetric_3}, and \autoref{fig:fewshotmetric_4}
% (Fig.~5, 15, 16, and 17 in the main paper)
report comprehensive few-shot scaling results for all subject pairs, covering transfer directions
$1\!\rightarrow\!\{2,5,7\}$,
$2\!\rightarrow\!\{1,5,7\}$,
$5\!\rightarrow\!\{1,2,7\}$, and
$7\!\rightarrow\!\{1,2,5\}$.

\paragraph{Consistent monotonic improvements.}
Across all transfer directions, we observe a clear and largely monotonic improvement as the number of few-shot anchor samples increases.
This trend is consistent across
low-level image fidelity metrics (PixCorr, SSIM, AlexNet(2/5)),
high-level semantic similarity metrics (Inception, CLIP, EffNet-B, SwAV),
as well as retrieval-based alignment metrics (fwd\_corr and bwd\_corr).
The smooth scaling behavior indicates that MindAdapter effectively leverages additional paired supervision to refine subject-specific residual corrections, rather than overfitting to a small anchor set.

% \paragraph{Strong performance in the low-shot regime.}
Notably, substantial performance gains are already achieved with as few as 32--64 shared stimuli.
In this low-shot regime, MindAdapter consistently outperforms the few-shot variant of MindAligner across nearly all evaluated metrics.
This observation highlights the strong data efficiency of the proposed residual calibration mechanism, demonstrating that a small number of topological anchors is sufficient to correct dominant cross-subject misalignment when combined with unpaired semantic regularization.

% \paragraph{Stability across transfer directions.}
Although different subject pairs exhibit varying baseline difficulty due to inter-subject variability, the overall scaling trends remain highly consistent.
For all source--target configurations, performance curves exhibit rapid early improvement followed by gradual saturation as the number of shots increases.
This stability suggests that MindAdapter does not rely on a specific alignment direction, but instead provides a robust and transferable calibration strategy applicable across diverse cross-subject mappings.

% \paragraph{Behavior of semantic and retrieval metrics.}
We further observe that semantic similarity metrics (e.g., CLIP, Inception) and retrieval accuracies (fwd\_corr, bwd\_corr) tend to saturate earlier than some low-level image fidelity metrics.
This behavior aligns with the design of the topology-anchored dual-stream objective, where semantic consistency under a frozen visual decoder constrains the residual adapter to correct visually salient errors first.
As more anchor samples are introduced, fine-grained structural details continue to improve, while global semantic alignment remains stable.
% \paragraph{Summary.}
Taken together, the results in \autoref{fig:fewshotmetric_1}, \autoref{fig:fewshotmetric_2}, \autoref{fig:fewshotmetric_3}, and \autoref{fig:fewshotmetric_4} demonstrate that MindAdapter exhibits reliable, monotonic, and direction-agnostic performance scaling with increasing few-shot supervision.
These findings further validate the effectiveness of the proposed coarse-to-fine residual calibration framework and its suitability for practical cross-subject brain decoding under realistic data constraints.

\begin{figure*}[t!]
\centering
\includegraphics[width=0.46\textwidth]{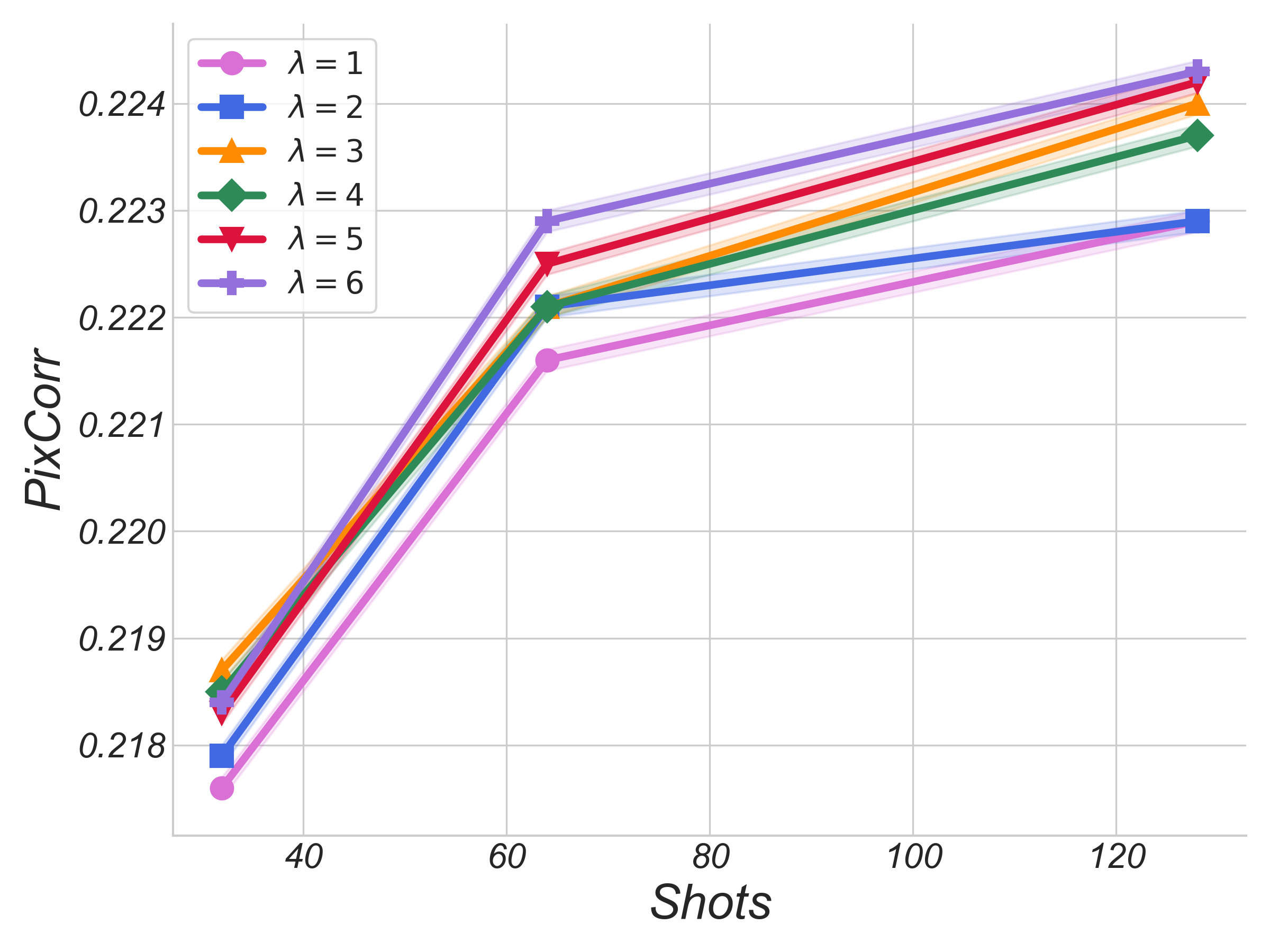}
% \vspace{-1.5em}
\caption{
Sensitivity analysis of the anchor–semantic loss balancing weight. Each curve corresponds to a different weight setting, showing PixCorr performance under varying numbers of shots.
} 
\label{fig:dualstreamparas}
\end{figure*}

\begin{table*}[!t]
\centering
\caption{
Quantitative comparison between MindAligner and MindAdapter under the 32-shot calibration setting across multiple target subjects, evaluated by low-level image fidelity (PixCorr, SSIM), perceptual similarity (AlexNet-2/5, Inception), and semantic alignment (CLIP, EffNet-B, SwAV). MindAdapter consistently improves reconstruction quality and semantic consistency over the few-shot-adapted MindAligner.
}
\resizebox{0.8\textwidth}{!}{
\begin{tabular}{@{}lcccccccc@{}}
\toprule
\textbf{Method} 
& \textbf{PixCorr$\uparrow$} 
& \textbf{SSIM$\uparrow$} 
& \textbf{Alex(2)$\uparrow$} 
& \textbf{Alex(5)$\uparrow$} 
& \textbf{Incep$\uparrow$} 
& \textbf{CLIP$\uparrow$} 
& \textbf{Eff$\downarrow$} 
& \textbf{SwAV$\downarrow$} \\
\midrule
MindAligner(subj1)
& 0.2136 & 0.4136 & 0.8764 & 0.9255 & 0.8319 & 0.8160 & 0.8004 & 0.4590 \\
Ours(subj1)
& \textbf{0.2231} & \textbf{0.4171} & \textbf{0.8783} & \textbf{0.9304} & \textbf{0.8340} & \textbf{0.8209} & \textbf{0.7920} & \textbf{0.4533} \\
\midrule
MindAligner(subj2)
& \textbf{0.2196} & \textbf{0.4274} & \textbf{0.8839} & \textbf{0.9393} & \textbf{0.8394} & \textbf{0.8312} & 0.7900 & 0.4486 \\
Ours(subj2)
& 0.2160 & 0.4265 & 0.8819 & 0.9366 & 0.8361 & 0.8307 & \textbf{0.7887} & \textbf{0.4480} \\
\midrule
MindAligner(subj5)
& 0.1936 & 0.4083 & \textbf{0.8511} & \textbf{0.9159} & \textbf{0.8469} & \textbf{0.8327} & \textbf{0.7776} & \textbf{0.4488} \\
Ours(subj5)
& \textbf{0.1998} & \textbf{0.4176} & 0.8485 & 0.9144 & 0.8420 & 0.8285 & 0.7833 & 0.4507 \\
\midrule
MindAligner(subj7)
& 0.1702 & 0.4069 & 0.8149 & \textbf{0.8809} & 0.7946 & 0.7671 & 0.8356 & \textbf{0.4864} \\
Ours(subj7)
& \textbf{0.1818} & \textbf{0.4117} & \textbf{0.8184} & 0.8794 & \textbf{0.7953} & \textbf{0.7711} & \textbf{0.8307} & 0.4871 \\
\bottomrule
\end{tabular}
}
\label{tab:32shot_comparison}
% \vspace{-3mm}
\end{table*}

\begin{figure*}[t!]
\centering
\includegraphics[width=0.8\textwidth]{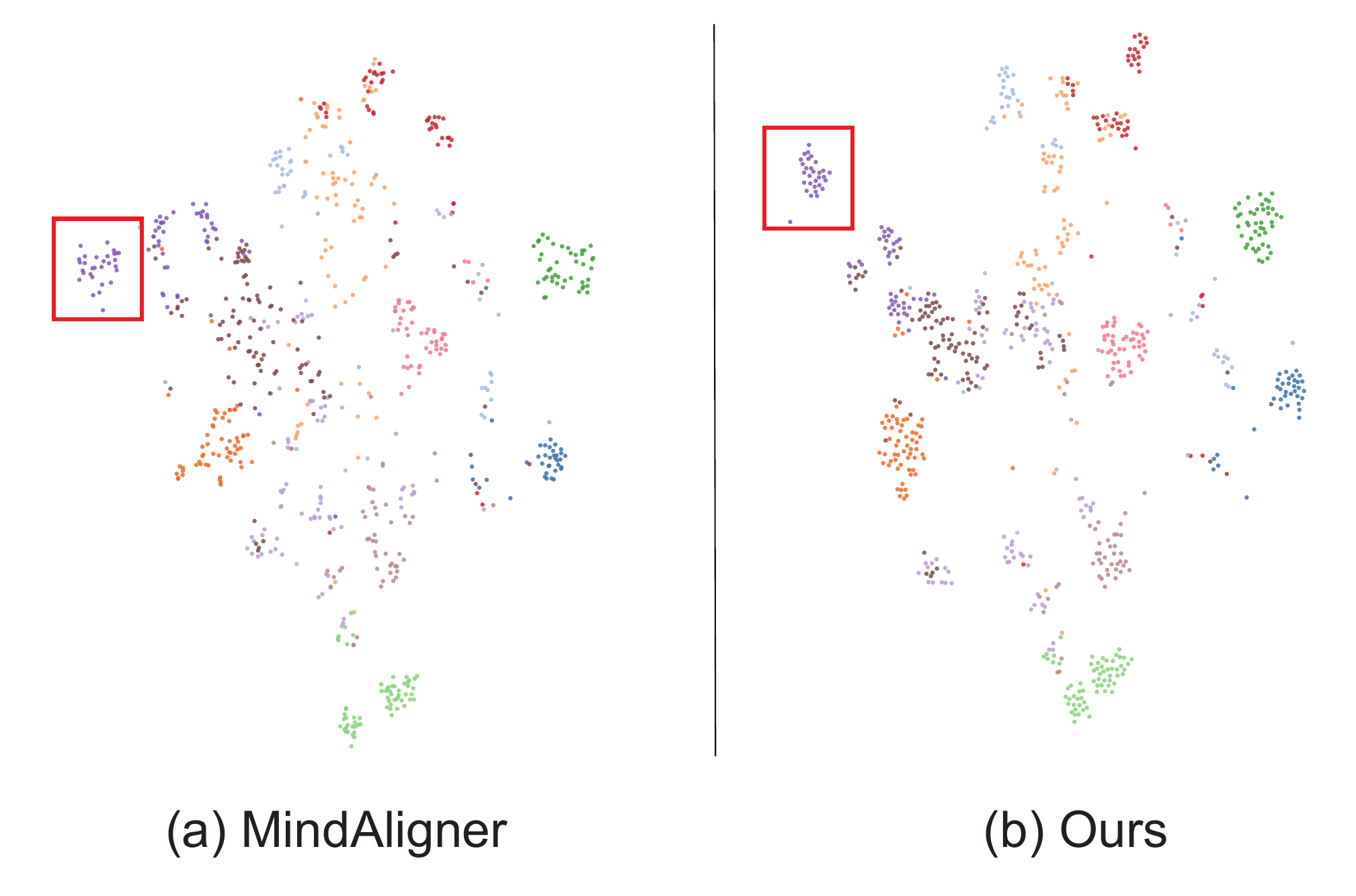}
% \vspace{-1.5em}
\caption{
t-SNE visualization of embedding spaces before and after calibration.
Compared to MindAligner, ours produces more compact and better-separated semantic clusters (highlighted), indicating improved cross-subject alignment and reduced representation dispersion after calibration.
} 
\label{fig:tsne}
% \vspace{-1.5em}
\end{figure*}

\begin{figure*}[t!]
\centering
\includegraphics[width=\textwidth]{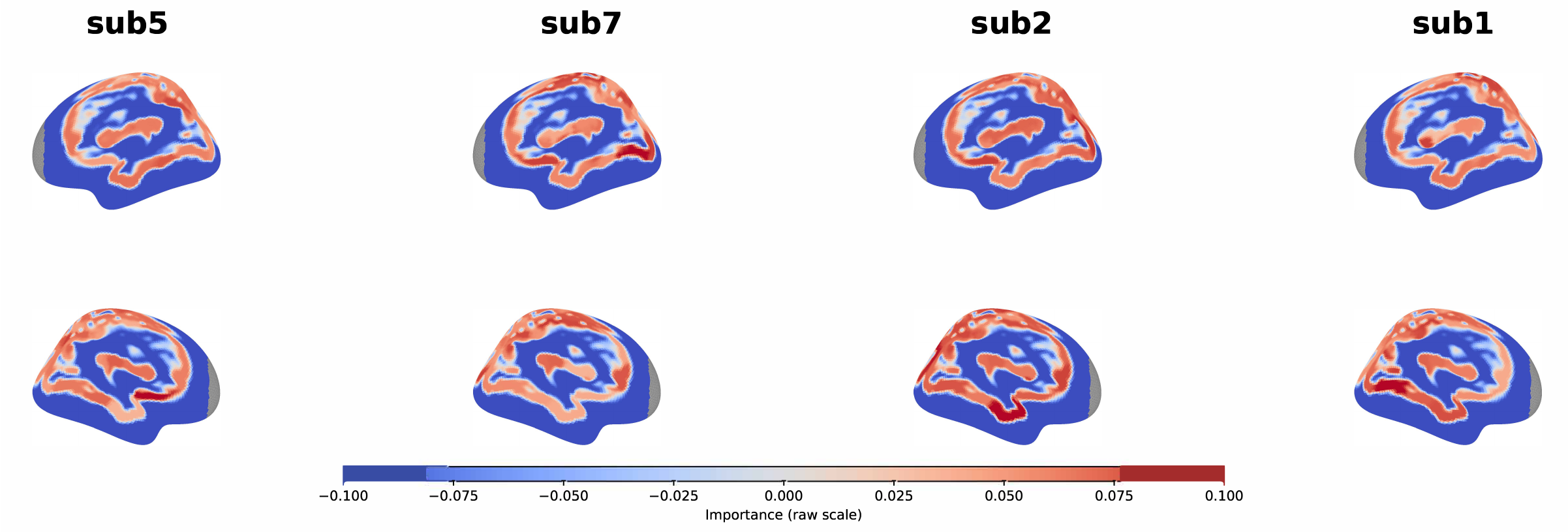}
% \vspace{-1.5em}
\caption{
% Surface‑based comparison of ROI‑level importance across four subject‑specific models (sub5, sub7, sub2, sub1), projected onto the Schaefer‑100 atlas on the fsaverage5 surface. The maps visualize relative differences in model‑derived importance on a shared cortical shell for qualitative comparison. 
Consistent model-derived functional organization across subjects.
Surface-based ROI-level importance maps derived from four calibrated subject-specific models and projected onto a common cortical surface. Despite subject-specific calibration, the learned importance patterns exhibit strong spatial consistency, indicating that MindAdapter preserves shared functional organization while injecting individualized residual corrections.
% Because a true subject‑specific voxel‑to‑ROI mapping was unavailable, these projections are for relative visualization only and should not be interpreted as anatomically localized effects.
} 
\label{fig:roi_vis}
% \vspace{-1.5em}
\end{figure*}

\section{Detailed Analysis of 32-Shot Quantitative Results}
\label{app:table3_analysis}

\autoref{tab:32shot_comparison} presents a quantitative comparison between the original MindAligner and the proposed MindAdapter under the 32-shot calibration setting across multiple target subjects.
MindAdapter performs few-shot calibration by attaching a lightweight nonlinear residual adapter on top of the frozen MindAligner backbone.

% \paragraph{Overall performance trends.}
Across all evaluated target subjects, MindAdapter consistently improves upon MindAligner on most low-level image fidelity metrics (PixCorr, SSIM).
These improvements indicate that even when a coarse linear cross-subject alignment is already available, substantial residual misalignment remains, which can be effectively corrected through few-shot residual calibration.

% \paragraph{Perceptual and semantic improvements.}
MindAdapter also yields clear gains on perceptual similarity metrics such as AlexNet(2/5) and Inception accuracy for most subjects.
Furthermore, consistent improvements in CLIP similarity demonstrate that MindAdapter enhances alignment in the visual semantic space, validating the role of semantic regularization during calibration.
These results confirm that correcting subject-specific residuals leads to better semantic fidelity beyond voxel-level alignment.

% \paragraph{Subject-wise robustness.}
While the absolute performance varies across subjects due to inter-subject variability, MindAdapter exhibits a stable improvement trend across nearly all metrics and subjects.
Even in cases where MindAligner performs competitively on individual metrics, MindAdapter achieves more balanced gains across complementary measures, reflecting improved overall reconstruction quality rather than metric-specific overfitting.

% \paragraph{Implications.}
The results in \autoref{tab:32shot_comparison} highlight a fundamental limitation of purely linear cross-subject alignment: although a pretrained BTM captures coarse functional correspondence, it is insufficient to resolve fine-grained, subject-specific discrepancies.
MindAdapter addresses this gap by introducing a nonlinear residual correction that can be reliably learned from as few as 32 shared stimuli, leading to consistent improvements in both structural fidelity and semantic alignment without modifying the original cross-subject mapper.









\end{document}